\newif\ifijrr
\ijrrtrue

\ifijrr
    \documentclass[Afour,sageh,times]{sagej}
\else
    \documentclass[lettersize,journal]{IEEEtran}
    \usepackage{cite}
\fi
\usepackage{amsmath,amssymb}
\usepackage{algorithm}
\usepackage{algpseudocodex}
\usepackage{colortbl}
\usepackage{url}
\usepackage{graphicx}
\usepackage{balance}
\usepackage{siunitx}
\usepackage[colorlinks,bookmarksnumbered,linkcolor=orange,filecolor=orange,citecolor=orange,urlcolor=orange]{hyperref}
\ifijrr
    \let\cite\citep
\fi

\newcommand{\p}[1]{\smallskip \noindent \textbf{{#1}.}}
\newcommand{\eq}[1]{Equation~(\ref{eq:#1})}
\newcommand{\fig}[1]{Figure~\ref{fig:#1}}

\newcommand{\paperabstract}{%
Robots that learn from demonstrations should extrapolate their policy despite state perturbations such as sensor noise, scene variation, and compounding errors. Prior works use regularizers to limit the policy's sensitivity to any perturbation at each individual state. By contrast, our insight is that nearby states should respond to the same perturbation consistently: if moving an irrelevant object does not change the policy at one state, it should not at a nearby state. We therefore develop inner product regularization (InneR). We show that the inner product on action space induces an inner product on state space through the policy Jacobian (the Fisher information for stochastic policies), whose null space contains the perturbations with no first-order effect on the policy. Regularizing each inner product alone only shrinks the policy's sensitivity. Instead, InneR clusters states with aligned inner products, averages each cluster, and minimizes the smallest eigenvalues of that average so that nearby states share a common null space. InneR thus enforces consistent responses without limiting policy expressiveness, and can be appended to existing imitation learning pipelines for deterministic or stochastic policies. Our experiments compare InneR to behavior cloning and norm regularization methods. In simulated experiments, InneR makes learned policies more robust to irrelevant state dimensions and to shifts between training and test distributions, whereas the norm penalties often perform worse than behavior cloning. On a real robot arm performing three manipulation tasks, InneR raises the average success rate of a diffusion policy from 18\% to 57\% without additional data.
}

\newcommand{\longpaperabstract}{%
Robots that learn from demonstrations should extrapolate their policy despite state perturbations such as sensor noise, scene variation, and compounding errors.
One approach to this robustness is regularization, i.e., adding an inductive bias during training that shapes the learned policy. Prior regularizers consider each state on its own and limit how sensitive the policy is to any perturbation at that state. By contrast, we explore how the policy's output shifts when given the same perturbation at different states. Our insight is that, locally, the robot should respond to state perturbations in a consistent manner: e.g., if changing an irrelevant object position does not impact the policy at one state, that same change should not shift the policy output at a nearby state. Applying this insight, we develop inner product regularization (InneR). We theoretically show that the inner product on action space induces an inner product on state space through the policy Jacobian, and that the null space of this induced inner product governs which state perturbations have a first-order effect on the policy outputs; for stochastic policies, this inner product is the Fisher information of the action distribution with respect to state. Regularizing each inner product by itself only shrinks the policy's sensitivity, but by clustering states whose inner products are aligned, averaging the inner products within each cluster, and then minimizing the smallest eigenvalues of that average, InneR shapes the learned policy so that nearby states share a common null space. Overall, InneR enforces consistent responses to state perturbations without limiting policy expressiveness, and can be appended to existing imitation learning pipelines for deterministic or stochastic policies. Our experiments compare InneR to behavior cloning and norm regularization methods across deterministic and stochastic policies. In simulated experiments, InneR makes learned policies more robust to irrelevant state dimensions and shifts between training and test distributions, whereas the norm penalties often perform worse than behavior cloning. On a real robot arm performing three manipulation tasks, InneR raises the average success rate of a diffusion policy from $18\%$ to $57\%$ without additional data.
}

\begin{document}

\ifijrr
    \runninghead{Christie et al.}
    \title{Inner Product Regularization for Imitation Learning}
    \author{Benjamin A. Christie\affilnum{1}, Tianyu Qiu\affilnum{2}, Amy H. Choi\affilnum{2,3}, \\David Fridovich-Keil\affilnum{2}, and Dylan P. Losey\affilnum{1}}
    \affiliation{\kern-\scriptspace\affilnum{1}Virginia Tech, Department of Mechanical Engineering\\
    \affilnum{2}The University of Texas at Austin, Department of Aerospace Engineering and Engineering Mechanics\\
    \affilnum{3}Korea Advanced Institute of Science and Technology, Department of Industrial and Systems Engineering
    }
    \corrauth{Benjamin A. Christie, Department of Mechanical Engineering, Virginia Tech, Blacksburg, VA, 24060, USA.}
    \email{benc00@vt.edu}
    \begin{abstract}
    \longpaperabstract
    \end{abstract}
    \keywords{Imitation Learning, Robust Control, Robot Learning}
    \maketitle
\else
    \title{Inner Product Regularization for Imitation Learning}
    \author{Benjamin A. Christie, Tianyu Qiu, Amy H. Choi, David Fridovich-Keil, Dylan P. Losey}
    \maketitle
    \begin{abstract}
    \paperabstract
    \end{abstract}
\fi

\section{Introduction} \label{sec:intro}

\begin{figure*}[t]
    \centering
    \includegraphics[width=1\linewidth]{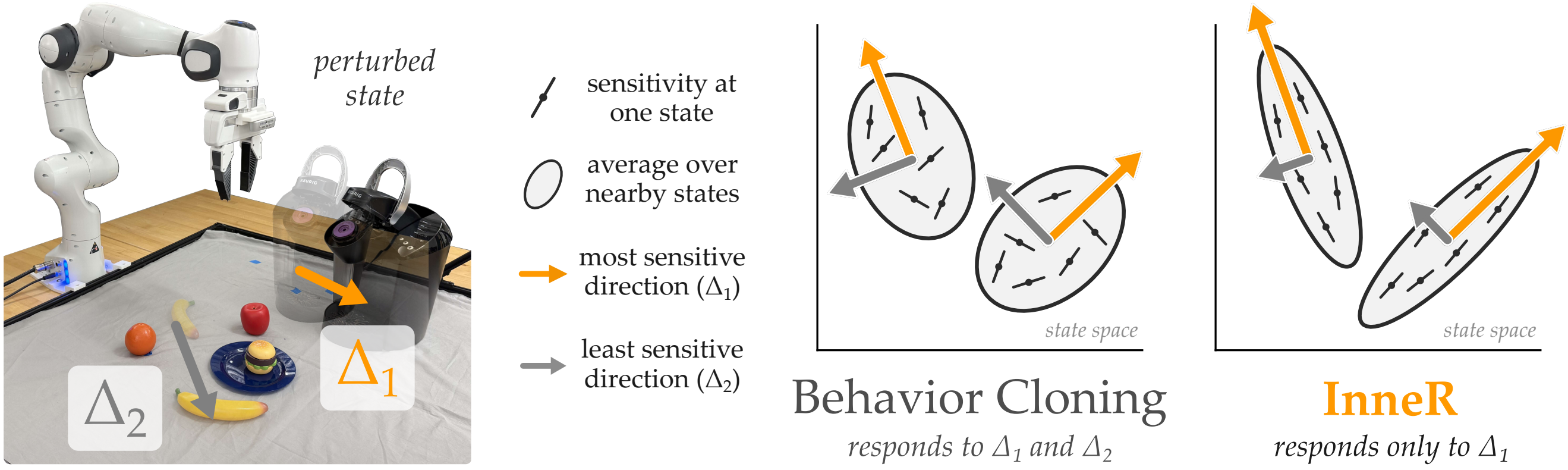}
    \caption{(Left) Robot learning to close a coffee machine. At test time the robot encounters a new state. 
    As compared to the states in its training dataset, this new state has some perturbations that are task-relevant (shifting the pose of the coffee machine, $\Delta_1$) and some that are irrelevant (shifting the banana, $\Delta_2$). (Right) Local state perturbations have a first-order effect on the policy outputs. At each state the policy is sensitive to some perturbation directions and ignores others; averaging over nearby states reveals the directions to which the policy is most and least sensitive. Under standard behavior cloning these responses are inconsistent across nearby states, and thus the policy reacts to both $\Delta_1$ and $\Delta_2$. We propose InneR, an inner product regularization approach that enforces similar responses to the same perturbation at nearby states: the policy still responds to $\Delta_1$ but now ignores $\Delta_2$.}
    \label{fig:front}
\end{figure*}

The goal of imitation learning is to extrapolate a generalized policy from a limited set of expert examples.
Consider \fig{front}: during training, the robot is shown how to close the coffee machine, and at test time it must generalize that policy across new scenarios, which expose it to states that are not in its training dataset \cite{spencer2021feedback}.
For example, sensor noise shifts the measured pose of objects, the scene itself varies between trials, and small errors in the robot's actions can compound over time.
Regularization techniques are often employed to enhance robustness in the face of these perturbations.
Instead of relying on additional data or expert oversight, regularization enforces an inductive bias on the learned policy during training.

Existing regularizers try to limit how sensitive the policy is to changes in state.
For instance, one standard approach is to minimize the norm of the model's parameter vector \cite{krogh1991simple, liu2021regularization}.
Other recent techniques penalize the norm of the policy Jacobian with respect to state \cite{hoffman2019robust, scarvelis2024nuclear}, or, for stochastic policies, the Fisher information of the action distribution with respect to input \cite{shen2019defending, gloeckler2023adversarial}.
These penalties share a common structure: each state is treated on its own, and the policy is made less sensitive to every perturbation at that state.
But not every perturbation should be ignored, e.g., a policy that is robust to moving clutter in the background should still respond when its target object shifts to the right.
Uniformly shrinking the policy's sensitivity dampens both responses, fundamentally limiting the robot's ability to control each of its action dimensions.

We propose an alternate perspective on regularization in behavior cloning.
Instead of asking how much the policy changes at a given state, we focus on how the policy responds to the \textit{same} perturbation at \textit{different} states.
Our intuition is that --- at least within a local neighborhood --- the robot should respond to perturbations in a consistent manner.
For example, if a shift in lighting does not affect the robot's policy at one state, the same shift should not affect the robot's policy at a nearby state.
Conversely, if a perturbation does substantially impact the task (e.g., the target object moves), then the robot's policy should change at both states.
This intuition implies structure, and therefore regularization, for the learned policy: 
\begin{center} \vspace{-0.25em}
    \textit{The first-order response to state perturbations should be locally consistent across the policy.}
    \vspace{-0.25em}
\end{center}
Applying this insight, we develop inner product regularization for imitation learning.
The designer starts by choosing an inner product on action space to measure how different two actions are.
We show that --- pulled back through the policy --- the inner product on action space induces an inner product matrix on state space, and this state-space inner product governs which local perturbations have a first-order effect on the policy outputs.
The same construction extends to stochastic policies, where the Fisher information of the action distribution with respect to state plays the role of the inner product matrix.
Perturbations in the null space of this matrix are ignored by the policy, while perturbations in its row space remain influential.
Our insight then becomes a condition on this null space: nearby states should share a common null space, making their responses to perturbations locally consistent.

We propose inner product regularization (\textbf{InneR}) as a tool for enforcing this local null space consistency.
InneR works alongside standard imitation learning algorithms by clustering the states in each training batch according to their inner product alignment, averaging the inner product matrices within each cluster, and then penalizing the shared null space.
When we average the inner product matrix across a set of states, the smallest eigenvalues of that average measure how far the states are from sharing a common null space.
Regularizing the policy to minimize these eigenvalues therefore encourages consistency, and this consistency does not come at the cost of expressiveness.
Since the regularizer is zero for any family of full-rank Jacobians with a common row space, InneR can align the directions that the policy ignores without reducing the policy's sensitivity to the directions that matter.

In summary, we make the following contributions:

\p{Connecting local perturbations to the inner product on state space}
We show that the inner product on action space induces an inner product matrix on state space through the policy Jacobian, and that the null space of this matrix characterizes which local perturbations have no first-order effect on the policy.
For stochastic policies, we show that this matrix is the Fisher information with respect to state.

\p{Aligning the null space using active subspace learning}
We show the smallest eigenvalues of the expected inner product are zero exactly when the individual inner products share a null space.
This yields an eigenvalue regularizer that enforces consistent responses to perturbations without reducing the rank of the policy. 
This method contrasts with commonly-employed Frobenius and nuclear norm penalties on the Jacobian.

\p{Localizing the regularization by inner product alignment}
We implement our theory with the InneR algorithm for deterministic or stochastic policies.
To encourage local consistency, InneR first measures the geometric alignment between inner product matrices.
InneR then clusters states which are already aligned, and regularizes each cluster separately so that the relevant state directions can change over the course of a task.

\p{Comparing to alternative regularizers}
We compare InneR to unregularized and regularized imitation learning frameworks across multiple model types in controlled simulations and in the real world.

\section{Related Work} \label{sec:related}

We develop a metric-based regularizer for imitation learning that shapes the state space inner product so that nearby states reject the same perturbations.

\p{Robustness in Imitation Learning}
A central challenge in imitation learning is \textit{covariate shift} \cite{spencer2021feedback}: at test time the robot encounters states outside of its training dataset, and small errors compound over the task \cite{ross2011reduction}.
Many approaches address this shift with more data, either by querying a human for corrections online \cite{ross2011reduction, kelly2019hg} or by injecting noise into the demonstrations \cite{laskey2017dart, qiu2026safety}.
Because policies readily latch onto task-irrelevant features \cite{dehaan2019causal}, a second line of work learns task-relevant state representations by backpropagating loss gradients \cite{parekh2026transmask} or applying image perturbations \cite{pari2021surprising}.
A third line treats the learner as a dynamical system and bounds the growth of errors with control theoretic tools \cite{mehta2025stable, pfrommer2022tasil, block2023provable}.
Our work requires neither additional data nor a latent representation: we regularize the policy directly in the state space it acts upon.

\p{Distance Metrics}
A core idea behind our approach is that a distance in action space induces a distance in state space: states that the policy maps to similar actions are functionally equivalent and therefore ``nearby.''
In reinforcement learning this idea underlies state abstractions that aggregate states with the same optimal action \cite{li2006towards} and bisimulation metrics that compare states through their rewards, transitions, or action distributions, and then shape a learned representation accordingly \cite{ferns2004metrics, zhang2020learning, agarwal2021contrastive}.
Recent research has also begun to leverage this idea in imitation learning, clustering states that lead to similar action sequences for robustness to changes in viewpoint and appearance \cite{lee2025class}.
Rather than learning a latent space, we regularize the pullback of the action space metric through the policy itself.
The same pullback appears in Riemannian motion policies \cite{ratliff2018riemannian} and in the geometry of deep generative models, which pull the output metric or its Fisher-Rao counterpart back through the decoder \cite{arvanitidis2017latent, beik2021learning, arvanitidis2022pulling}.
In these prior works the pullback metric is computed through a given map, whereas we shape that metric during training. 

\p{Jacobian Regularization}
In practice, our inner product shaping is a form of regularization on the policy \textit{Jacobian} with respect to state, whose singular values measure the policy's sensitivity to local perturbations.
Penalizing its magnitude dates back to double backpropagation \cite{drucker1992improving} and contractive autoencoders \cite{rifai2011contractive}, and remains standard in supervised learning \cite{hoffman2019robust}, reinforcement learning \cite{chen2025learning}, and imitation learning \cite{wu2025robust, xie2026learning}.
The magnitude is often measured by the Frobenius norm \cite{hoffman2019robust, xie2026learning} or by the nuclear norm \cite{scarvelis2024nuclear}.
For stochastic models, the analogous quantity is the Fisher information with respect to the input, whose trace or largest eigenvalue can be penalized directly \cite{shen2019defending, gloeckler2023adversarial}; Kullback-Leibler (KL)-based smoothness penalties regularize the same matrix implicitly \cite{miyato2018virtual, zhang2020robust}.
These penalties can be enforced locally, at individual states \cite{kobayashi2022l2c2, mysore2021regularizing}, or globally, through constraints on the network weights \cite{gouk2021regularisation}.
Closer to our insight, tangent propagation makes a model invariant along state directions that are specified by the designer \cite{simard1991tangent, ross2017right}.

Our approach fundamentally differs by targeting the expected inner product across neighboring states: norm penalties indiscriminately reduce the policy's sensitivity and rank, limiting the robot's ability to control independent action dimensions, whereas our method preserves action rank and learns which directions to ignore.
This matrix is the vector-valued analogue of the expected gradient outer product, whose leading eigenvectors span the input directions a learned function depends on \cite{radhakrishnan2024mechanism} and which active subspace methods use for \textit{post-hoc} dimensionality reduction \cite{constantine2014active, zahm2020gradient}.
We instead adapt this geometric framework into a training regularizer that penalizes the trailing eigenvalues, enforcing a perturbation-rejecting null space in the policy Jacobian at neighboring states.

\section{Problem Statement} \label{sec:problem}

We consider a standard imitation learning problem where a robot tries to learn a control policy from a dataset of offline expert demonstrations.

\p{States and Actions}
Let $\mathcal{S}$ and $\mathcal{A}$ denote the state and action spaces, with states $s \in \mathcal{S}$ and actions $a \in \mathcal{A}$.
We focus on continuous environments where $\mathcal{S} \subseteq \mathbb{R}^n$ and $\mathcal{A} \subseteq \mathbb{R}^m$ are subsets of finite-dimensional inner product spaces.
Furthermore, we assume that the state dimension exceeds the action dimension ($n > m$).
In our experiments the robot has direct access to a state vector (e.g., we do not consider image space observations); the state could be measured using object detection, segmentation, or pose estimation methods \cite{redmon2016you, carion2026sam, wen2024foundationpose}.
Our actions consist of robot position or velocity commands.

\p{Dataset}
The robot learns from an offline set of expert task demonstrations.
Specifically, the robot is given a dataset $\mathcal{D} = \{(s_1, a_1), \ldots, (s_K, a_K)\}$ of states and the corresponding actions to take at those states.
We let $p(s)$ denote the distribution of states visited by the expert, so that the states in $\mathcal{D}$ are samples from distribution $p(s)$.

\p{Policy and Loss Function}
Based on the dataset, the robot extrapolates a control policy $\pi_\theta : \mathcal{S} \rightarrow \mathcal{A}$ that maps from states to actions.
For now we will assume that this policy is deterministic (Section~\ref{sec:determ}), but later we will also explore stochastic policy variants (Section~\ref{sec:stoch}).
The control policy is instantiated as a differentiable model (e.g., a neural network) parameterized by weights $\theta$.
We assume that $\pi_\theta$ is defined on an open set containing $\mathcal{S}$, and that the policy is differentiable with respect to both its input $s$ and weights $\theta$.
We learn these weights to minimize a loss function $\mathcal{L}(\theta)$ across the dataset.
For example, in behavior cloning the deterministic policy is trained to match the actions in the dataset:
\begin{equation} \label{eq:P1}
    \mathcal{L}(\theta) = \frac{1}{K} \sum_{(s, a) \in \mathcal{D}} \|a - \pi_\theta(s)\|^2
\end{equation}
Our approach is agnostic to the specific choice of this loss function.
We simply assume that the robot trains $\pi_\theta$ to minimize loss $\mathcal{L}$ across the given dataset $\mathcal{D}$.

\p{Perturbations}
For any state $s$ in the dataset $\mathcal{D}$ the robot knows exactly what action it should take.
But at test time the robot will inevitably experience new states $s' = s + \Delta$, where perturbation $\Delta \in \mathbb{R}^n$ represents a shift due to sensor noise, environmental variations, compounding execution errors, or unseen environment configurations.
We specifically consider local perturbations $\|\Delta\| < \epsilon$ where the state is slightly modified; here $\|\cdot\|$ is the Euclidean norm and $\epsilon > 0$ is small enough that a first-order Taylor expansion of $\pi_\theta$ around $s$ is accurate.
Our goal is to ensure that the policy is robust to these local perturbations. Put another way, the robot should not change its actions if a state perturbation is irrelevant to the current task, but it should remain sensitive to task-relevant changes to the state.
\section{Inner Product Regularization: \\ \textbf{Deterministic Policies}} \label{sec:determ}

Our insight is that --- if a given perturbation $\Delta$ has no impact on the policy at state $s_1$ --- that same perturbation should not impact the policy at a nearby state $s_2$.
We can therefore improve robustness to local perturbations by regularizing the learned policy $\pi_\theta$ to enforce this spatial consistency.
Here we formalize policy responses to local perturbations from a geometric perspective.
We use the inner product on action space to induce an inner product matrix on state space: this inner product matrix determines what local perturbations have a first-order effect on the policy output (Section~\ref{sec:D1}).
Next, we apply principles of active subspace learning to shape the null space of the expected inner product across the state space (Section~\ref{sec:D2}).
We reach an eigenvalue condition that, when satisfied, causes the policy to reject similar perturbations at nearby states to the first order.

\subsection{Inner Products in Action and State Spaces} \label{sec:D1}

Let $a$ be the policy's predicted action for the current task at state $s$, and let $a'$ be the policy's predicted action for the current task at a new state $s' = s + \Delta$.
We want to know whether perturbation $\Delta$ has a meaningful impact on the policy's behavior---i.e., does adding $\Delta$ cause the model's predicted action to change?

\p{Inner Product on Action Space}
To answer this question, we first need to know the difference between actions $a$ and $a'$.
We therefore define a metric $d_a : \mathcal{A} \times \mathcal{A} \rightarrow \mathbb{R}$ on the action space.
We choose this metric to be induced by an inner product on $\mathbb{R}^m$, so that it takes the form:
\begin{equation} \label{eq:J1}
    d_a(a', a) = \| a' - a \|_M = \sqrt{(a'-a)^T M (a'-a)}
\end{equation}
where $M \in \mathbb{R}^{m\times m}$ is the symmetric and positive definite action inner product matrix.
Often designers set $M$ as the identity matrix; but one can also tune $M$ to alter the relative sensitivity of different action dimensions, so that small changes cause a large increase in distance $d_a$.

\p{Induced Distance in State Space}
Metric $d_a$ quantifies the distance in action space between $a$ and $a'$.
Next, we apply this metric to understand the distance in state space between $s$ and $s' = s + \Delta$.
We define $d_s$ as the pullback pseudometric induced on the state space by $d_a$ via the policy $\pi_\theta$:
\begin{equation} \label{eq:J2}
    d_s(s', s) = d_a\big(\pi_\theta(s'), \pi_\theta(s)\big)
\end{equation}
Under pseudometric $d_s$ states are close together if they map to similar actions, and states are far apart if they map to dissimilar actions.
Note that this distance in state space is inherently task-specific: as the policy $\pi_\theta$ changes for different tasks, different states become closer together or farther apart.
Put another way, the distance pseudometric $d_s$ is a function of both the chosen inner product on action space $M$ as well as the task-specific policy $\pi_\theta$ the robot is learning.

Because $d_a$ is a metric, by definition the induced $d_s$ satisfies non-negativity, symmetry, and the triangle inequality.
However, $d_s$ fails the identity property because the policy maps a continuous higher-dimensional state space into a lower-dimensional action space ($n > m$), and therefore $\pi_\theta$ is inherently non-injective.
We conclude that $d_s$ is a pseudometric with $d_s(s', s) = 0$ for some $s' \neq s$, and therefore there exist some perturbations $\Delta \neq 0$ that do not change the resulting action, i.e., $\pi_\theta(s+\Delta) = \pi_\theta(s)$.

\p{Inner Product on State Space}
To determine which perturbations $\Delta$ do not impact the policy's behavior, we focus on the pseudometric $d_s$.
Starting with the expression $a' - a$ and applying a first order Taylor Series approximation to the robot policy, we obtain:
\begin{equation} \label{eq:J3}
    a' - a = \pi_\theta(s') - \pi_\theta(s) \approx J(s) \Delta
\end{equation}
where perturbation $\Delta = s' - s$ and $J(s) \in \mathbb{R}^{m\times n}$ is the Jacobian of the policy with respect to state evaluated at a given state $s$:
\begin{equation} \label{eq:J4}
    J(s) = \frac{\partial \pi_\theta}{\partial s} \bigg|_{s}
\end{equation}
We note that the first order approximation in \eq{J3} is increasingly accurate when $\|\Delta\| \rightarrow 0$ and the perturbations become increasingly small.

Plugging \eq{J3} back into Equations~(\ref{eq:J1}) and (\ref{eq:J2}), we obtain a local approximation of the distance metric:
\begin{equation} \label{eq:J5}
    d_s(s', s) \approx \| J(s)\Delta\|_M = \|\Delta\|_{A(s)}
\end{equation}
Here $A \in \mathbb{R}^{n \times n}$ is the symmetric and positive semi-definite inner product matrix on the state space:
\begin{equation} \label{eq:J6}
    A(s) = J(s)^T M J(s)
\end{equation}
Strictly speaking, because $A(s)$ is only positive semi-definite, it defines a degenerate inner product (and $\|\cdot\|_{A(s)}$ is a seminorm).
Recalling that $M$ is the constant inner product matrix on the action space, $A$ is now the induced inner product matrix on the state space, and the mapping between these two inner products is determined by the policy Jacobian $J$ at a given state $s$.
The inner product $A$ also reflects how distances in state space depend upon the task: when the robot learns a new policy $\pi_\theta$, it changes the Jacobian $J$, and this yields a different inner product matrix $A$.

\p{Interpreting the Inner Product}
The inner product matrix in state space determines how the policy reacts to perturbations.
By construction, $A$ has a null space of at least dimension $n - m$ (i.e., state dimension minus action dimension): because $M$ is positive definite, the null space of $A(s)$ equals the null space of $J(s)$, and $\operatorname{rank}(J(s)) \leq m$.
The null space has dimension exactly $n - m$ when $J(s)$ has full row rank $m$, and local perturbations $\Delta$ that lie in this null space have no first-order effect on the policy's actions.
Conversely, local perturbations $\Delta$ with a non-zero component in the row space of $A$ have a non-zero first-order effect on the policy outputs. 
Because $A$ depends on Jacobian $J$, these null and row spaces may vary across both the system state and the intended task.

\subsection{Active Subspace Learning} \label{sec:D2}

In \eq{J6} we presented the inner product matrix $A$ that determines which perturbations $\Delta$ have a first-order impact on the policy's behavior.
In what follows, we explore how to regularize $A$ so that the policy is robust to irrelevant perturbations while still generalizing task-relevant actions to previously unseen states.
Our approach draws from active subspace learning \cite{constantine2014active}, a technique that seeks to identify and reject the input directions which have a negligible impact on the model's output.

\p{Connecting the Inner Product to Policy Robustness}
To understand how to regularize $A$, we apply our core insight: if a perturbation $\Delta$ does not impact the policy output at state $s$, that same perturbation should not impact the policy output at states near $s$.
Written in terms of our metric on state space, for every perturbation $\Delta \in \mathbb{R}^n$:
\begin{equation} \label{eq:J7}
    \| \Delta \|_{A(s_1)} = 0 \implies \| \Delta \|_{A(s_2)} = 0
\end{equation}
where $s_1$ and $s_2$ are nearby states, e.g., states within some region $\mathcal{B}$.
We will formalize these local regions in Section~\ref{sec:algorithm}.
For now --- to better introduce our regularization concept --- we simply treat $s_1$ and $s_2$ as arbitrary states in $\mathcal{S}$.

One naive method for satisfying \eq{J7} is to constrain $A$ to be constant across $s \in \mathcal{S}$.
If $A(s_1) = A(s_2) ~\forall s_1, s_2 \in \mathcal{S}$, then the policy has a consistent response to perturbations.
But by forcing $A$ to be invariant to state, we also constrain the policy to have an identical Jacobian $J$ across $\mathcal{S}$ (up to an isometry in the action space), and this prevents the policy $\pi_\theta$ from learning nuanced behaviors.
Rather than regularizing the entire inner product $A$, we instead focus on its null space, since local perturbations $\Delta$ which lie in this null space have no first-order effects on the policy.
More specifically, \eq{J7} holds for every perturbation $\Delta$ exactly when the null spaces of both $A(s_1)$ and $A(s_2)$ coincide.
We can therefore enforce our core insight and make nearby states robust to similar perturbations by regularizing $A$ to maintain a consistent null space across subsets of the state space.

\p{Expected Inner Product}
In what follows we outline how to perform this null space regularization on the state space inner product matrix.
For clarity of exposition, here we will assume that we want to enforce a consistent null space across all the states $s$ in dataset $\mathcal{D}$.
To do so, we start averaging the inner product matrix across these individual states:
\begin{equation} \label{eq:J8}
    \bar{A} = \mathbb{E}_{s \sim p(s)} \big[A(s)\big] \approx \frac{1}{K} \sum_{(s, a) \in \mathcal{D}} A(s)
\end{equation}
By construction, $\bar{A}$ is symmetric positive semi-definite.
Because each $A(s)$ is also positive semi-definite, $\Delta^T \bar{A} \Delta = 0$ if and only if $\Delta^T A(s) \Delta = 0$ for every state in the average.
Hence the null space of $\bar{A}$ is exactly the intersection of the individual null spaces, and the rank of $\bar{A}$ grows as these null spaces become less aligned; if they share no common direction, $\bar{A}$ is strictly positive definite.

\p{Eigenvalues of the Expected Inner Product}
We next extract the eigenvalues of $\bar{A}$ sorted from largest to smallest: 
\begin{equation} \label{eq:J9}
    \operatorname{eig}(\bar{A}) = (\lambda_1, \ldots, \lambda_m, \lambda_{m+1}, \ldots, \lambda_n)
\end{equation}
Because there are $m$ action dimensions, the eigenvectors associated with the first $m$ eigenvalues span the active subspace, i.e., the state directions to which the policy is most sensitive on average; these eigenvalues are not part of our regularization.
By contrast, the smallest $n-m$ eigenvalues measure how far the individual null spaces are from sharing a common $(n-m)$-dimensional subspace.
When $\lambda_{m+1} = \cdots = \lambda_n = 0$, then all the inner products $A(s)$ must share an identical null space and \eq{J7} is satisfied.

This result provides our main theoretical takeaway.
To shape the imitation learning policy $\pi_\theta$ so that it is consistently robust to the same state perturbations, we can compute the expected inner product matrix across (sections of) the state space, and then regularize the policy during training to minimize the $n-m$ smallest eigenvalues of that expected inner product matrix.
Because the null space and row space of $J(s)$ are orthogonal complements, sharing an $(n-m)$-dimensional null space is equivalent to sharing an $m$-dimensional row space.
Accordingly, the policy $\pi_\theta$ can still change how it maps the relevant state directions to actions as the state varies, but the set of irrelevant state directions remains fixed.

\p{Connection to Jacobian Regularization}
This construction has clear connections to existing Jacobian regularization techniques.
Consider \eq{J6}: because $M$ is a symmetric positive definite matrix, the null space of matrix $A$ is identical to the null space of Jacobian $J$.
Methods that regularize $J$ or the null space of $J$ may therefore also regularize the null space of the inner product matrix.
To better connect this Jacobian regularization with our inner product approach, below we will assume that $M=I$.
Existing works regularize either the Frobenius or nuclear norm of the policy Jacobian:
\begin{itemize}
    \item \textit{Frobenius Norm} $\|J\|_F$. Regularizing the squared Frobenius norm of $J(s)$ is equivalent to minimizing the sum of the eigenvalues of $A(s)$. In other words, $\|J\|^2_F$ is equivalent to the nuclear norm $\|A\|_*$.
    \item \textit{Nuclear Norm} $\|J\|_*$. Regularizing the nuclear norm of $J(s)$ is equivalent to minimizing the sum of the square roots of the eigenvalues of $A(s)$. This increasingly pushes the smaller eigenvalues of $A$ towards zero as compared to the squared Frobenius norm. 
\end{itemize}
Note that each individual policy Jacobian $J(s) \in \mathbb{R}^{m \times n}$ has at most rank $m$ and a null space of at least dimension $n-m$.
This null space already exists, and neither regularizer aligns it across states.
The squared Frobenius norm shrinks every singular value of $J(s)$ in proportion to its size, uniformly reducing the policy's sensitivity in all directions, including task-relevant ones.
The nuclear norm regularizer drives smaller singular values of $J(s)$ to zero more strongly.
However, neither outcome is necessarily desirable.
We typically want the rank of $J$ to be $m$, since this means that changing the state can separately affect each individual action dimension (e.g., the different joint or gripper positions).
When we increase the dimension of the null space of individual $J(s)$ matrices (e.g., through the nuclear norm), we are actually decreasing the rank of $A(s)$, and limiting how meaningful perturbations can guide the policy output.
By contrast, our regularizer only penalizes the components of $J(s)$ that act on a shared subspace: it is zero for any family of full-rank Jacobians with a common row space, and therefore does not require reducing rank.

At a higher level, applying regularization to individual $J(s)$ matrices does not encourage consistent responses to state perturbations.
The policy $\pi_\theta$ may have completely different responses to the same perturbation $\Delta$ at nearby states $s_1$ and $s_2$, and separately regularizing $J(s_1)$ and $J(s_2)$ does not penalize this.
The eigenvalues $\lambda_{m+1}, \ldots, \lambda_n$ in \eq{J9} can only emerge with the expected inner product matrix $\bar{A}$ (the equivalent eigenvalues are all zero for $A(s)$), and this fundamentally separates our methodology from approaches that regularize individual policy Jacobians.

\section{Inner Product Regularization: \\ \textbf{Stochastic Policies}} \label{sec:stoch}

Thus far our proposed inner product regularization has explored deterministic policies of the form $a = \pi_\theta(s)$.
However, state of the art imitation learning policies are often \textit{stochastic} \cite{urain2025survey}.
We will specifically focus on Gaussian policies, normalizing flows, and diffusion policies.
Instead of outputting a single action for a given state, these stochastic policies define a state-dependent probability distribution over actions:
\begin{equation} \label{eq:S1}
    a \sim \pi_\theta(\cdot \mid s)
\end{equation}
Just like their deterministic counterparts, stochastic policies should be robust to local perturbations.
Consider two nearby states $s_1$ and $s_2$ as well as a perturbation $\Delta$.
If perturbation $\Delta$ has no impact on the policy's action distribution at state $s_1$, then that same perturbation should not change the policy's action distribution at state $s_2$.

Our strategy for regularizing stochastic policies parallels the geometric analysis from Section~\ref{sec:determ}.
We will use divergence in the action distribution to induce an inner product matrix on state space, average that inner product across the dataset, and then regularize its smallest eigenvalues to enforce a shared null space.
The key difference is how the inner product is computed.
In Section~\ref{sec:S1} we show that the stochastic inner product in state space is actually the Fisher information matrix $F(s)$.
To better connect this result to $A(s)$, and to move towards practical implementation, we work through examples with Gaussian and diffusion policies in Section~\ref{sec:S2}.

\subsection{The Fisher Information Matrix as an Inner Product} \label{sec:S1}

We begin by formalizing the distance between two action distributions.
Let $s$ be the current state, and let $s' = s + \Delta$ be a nearby state resulting from perturbation $\Delta$.
To quantify the difference between the action distributions produced by the policy at $s$ and $s'$, we use Kullback-Leibler (KL) divergence:
\begin{multline} \label{eq:S2}
    d_s(s',s) = D_{KL}\big(\pi_\theta(\cdot \mid s) \,\|\, \pi_\theta(\cdot \mid s')\big) =  \\ \mathbb{E}_{a \sim \pi_\theta(\cdot \mid s)}\big[\log{\pi_\theta(a \mid s)} - \log{\pi_\theta(a \mid s')} \big]
\end{multline}
Like \eq{J2}, KL divergence asserts that two states are close together if they map to similar action distributions, and far apart if their action distributions diverge.
Unlike \eq{J2}, $d_s$ here is not a pseudometric: it is not symmetric, but both directions agree to second order and $d_s$ behaves locally like a squared distance as we show in \eq{S6}.

Although KL divergence itself is not a metric, we can obtain a squared distance on state space by applying a Taylor series approximation to \eq{S2}.
Define $g_\theta(a,s)$ as the gradient of the log probability of an action $a$ with respect to state $s$:
\begin{equation} \label{eq:S3}
    g_\theta(a,s) = \nabla_s \log{\pi_\theta(a \mid s)} \in \mathbb{R}^{n}
\end{equation}
and let $H_\theta(a,s) = \nabla^2_s \log{\pi_\theta(a \mid s)} \in \mathbb{R}^{n \times n}$ be the corresponding Hessian.
Expanding $\log{\pi_\theta(a \mid s')}$ in \eq{S2} to second order around $s' = s$, the approximation becomes:
\begin{multline} \label{eq:S4}
    d_s(s',s) \approx -\mathbb{E}_{a \sim \pi_\theta(\cdot \mid s)}\left[g_\theta(a,s)^T\right] \Delta \\
    -\frac{1}{2}\Delta^T \mathbb{E}_{a \sim \pi_\theta(\cdot \mid s)}\left[H_\theta(a,s)\right] \Delta
\end{multline}
We can further simplify \eq{S4} by recognizing that the first order term is zero due to the fact that $\Delta = 0$ is a global minimizer of $d_s$.
The second order term can be rewritten with the Fisher information identity, $\mathbb{E}_{a \sim \pi_\theta(\cdot \mid s)}[H_\theta(a,s)] = -\mathbb{E}_{a \sim \pi_\theta(\cdot \mid s)}[g_\theta(a,s) \, g_\theta(a,s)^T]$.
This leaves us with:
\begin{equation} \label{eq:S5}
    d_s(s',s) \approx \frac{1}{2}\Delta^T \mathbb{E}_{a \sim \pi_\theta(\cdot \mid s)}\left[g_\theta(a,s) \, g_\theta(a,s)^T\right] \Delta
\end{equation}
Interestingly, the Hessian of the KL divergence with respect to the perturbation $\Delta$ is the Fisher information matrix $F(s)$ of the action distribution with respect to the state $s$.
This version of the Fisher information matrix treats $s$ as the parameter of the distribution, as opposed to the more common Fisher information with respect to the model weights $\theta$.
Matrix $F(s)$ quantifies how sensitive the policy's probability distribution is to changes in state $s$.
Rewriting \eq{S5}, $F(s)$ defines an inner product matrix on the state space for stochastic policies:
\begin{equation} \label{eq:S6}
    d_s(s', s) \approx \tfrac{1}{2}\|\Delta\|^2_{F(s)}
\end{equation}
\begin{equation} \label{eq:S7}
    F(s) =  \mathbb{E}_{a \sim \pi_\theta(\cdot \mid s)}\left[g_\theta(a,s) \, g_\theta(a,s)^T\right]
\end{equation}
where $F(s) \in \mathbb{R}^{n\times n}$ is a symmetric positive semi-definite matrix.
Comparing \eq{S6} to \eq{J5}, we see that $F(s)$ is analogous to $A(s)$.
Local state perturbations which lie in the null space of $A(s)$ do not change the policy's \textit{action} to the first order; similarly, state perturbations which lie in the null space of $F(s)$ do not change the policy's \textit{action distribution} to the first order.

\p{Expected Inner Product and Eigenvalues}
Given the analogy between $A(s)$ and $F(s)$, the next steps for active subspace regularization follow from Section~\ref{sec:D2}.
To ensure that nearby states are robust to the same perturbations, we first calculate an average inner product $\bar{F}$.
We then extract and minimize the $n-r$ smallest eigenvalues of $\bar{F}$, where $n > r \geq m$ is the estimated rank of $F(s)$.
The rank of $F(s)$ is often greater than action dimension $m$, since a state perturbation can change the shape of the action distribution (e.g., its covariance) and not only its mean.
Overestimating $r$ penalizes fewer eigenvalues than the shared null space should contain, which weakens the regularizer.
Underestimating $r$ demands more zero eigenvalues than desirable, causing the regularizer to penalize relevant state dimensions and compete with the imitation loss $\mathcal{L}$.
In our experiments we therefore choose conservatively ($r = 2m$); alternatively, $r$ can be inferred from the spectrum of $\bar{F}$ for a briefly trained policy.
We leave this adaptive estimate of $r$ for future work.

\p{Connection to Fisher Information Regularization}
Prior work has regularized the Fisher information with respect to the model input to make classifiers and conditional density estimators robust to adversarial perturbations, e.g., by suppressing its largest eigenvalue~\cite{shen2019defending} or penalizing its trace~\cite{gloeckler2023adversarial}.
Like the Jacobian penalties in Section~\ref{sec:D2}, these methods regularize each $F(s)$ on its own and lower the penalty by shrinking the policy's sensitivity.
InneR differs in two ways.
First, we are regularizing an averaged $\bar{F}$ instead of individual $F(s)$ matrices. 
Second, we target direction rather than magnitude: the loss in \eq{A5} keeps the policy highly sensitive to task-relevant perturbations. 

\subsection{Application to Gaussian and Diffusion Policies} \label{sec:S2}

Here we apply the analysis from Section~\ref{sec:S1} to Gaussian and diffusion policies.
We derive an exact or estimated Fisher information matrix, and show how our approach applies to these common imitation learning frameworks.

\p{Relating Stochastic and Deterministic}
We first study a Gaussian setting to better connect the state-space inner products for stochastic policies, $F(s)$, and deterministic policies, $A(s)$. 
Consider a policy of the form: $\pi_\theta(a \mid s) = \mathcal{N}(a; \mu_\theta(s), \Sigma)$.
Here $\mathcal{N}$ is a Gaussian distribution over actions with state-dependent mean $\mu_\theta(s)$ and constant covariance $\Sigma$.
Let $J(s) = \partial \mu_\theta / \partial s \in \mathbb{R}^{m \times n}$ be the Jacobian.
This Jacobian is equivalent to \eq{J4} where the deterministic policy output is replaced by the mean action.

Applying \eq{S3}, we have that: 
\begin{equation} \label{eq:S8}
    g_\theta(a,s) = J(s)^T \Sigma^{-1}(a - \mu_\theta(s))    
\end{equation}
Plugging this value for $g_\theta(a,s)$ into \eq{S7} and taking the expectation, the Fisher information matrix is:
\begin{equation} \label{eq:S9}
    F(s) = J(s)^T \Sigma^{-1} J(s) 
\end{equation}
Gaussian policies therefore provide a setting where the inner product can be exactly computed.
Comparing \eq{S9} with \eq{J6}, we find that --- for Gaussian policies with constant covariance --- the inner products are equivalent when $M = \Sigma^{-1}$.
Recall that matrix $M$ specifies the relative sensitivity of different action dimensions.
The inverse covariance $\Sigma^{-1}$ is functionally equivalent: relatively smaller or larger values in $\Sigma$ indicate which action dimensions should have a tighter distribution, and inverting this matrix makes \eq{S9} increasingly sensitive to changes in action dimensions with the least variability.
Hence, Gaussian policies provide an example of the direct relationship between the inner products $A(s)$ and $F(s)$.

\p{Diffusion Policies}
We next outline how we estimate the inner product matrix for diffusion policies \cite{chi2025diffusion}.
Here the policy $\pi_\theta$ predicts a chunk of $H$ future actions $\mathbf{a} \in \mathbb{R}^{H \times m}$ based on the last $n_0$ states.
Accordingly, the policy input is the stacked state sequence $s \in \mathbb{R}^n$ with $n = n_0 n_s$, where $n_s$ is the dimension of a single state.
Unlike Gaussian policies, a diffusion policy does not provide its likelihood $\pi(\mathbf{a} \mid s)$ in closed form, and thus we cannot directly compute the gradient $g_\theta$ from \eq{S3}.

In our experiments we approximate $g_\theta$ using the evidence lower bound (ELBO) of the policy's reverse diffusion process.
Let $\ell_\theta(\mathbf{a}, s; t, \varepsilon) = \|\varepsilon - \varepsilon_\theta(\sqrt{\bar\alpha_t}\,\mathbf{a} + \sqrt{1-\bar\alpha_t}\,\varepsilon,\, t,\, s)\|^2$ be the denoising loss, where $t \sim \mathcal{U}\{1, \dots, T\}$ is a diffusion step, $\bar{\alpha}$ is the cumulative noise retention parameter, and $\varepsilon \sim \mathcal{N}(0, I)$ is the added noise \cite{ho2020denoising}.
The negative ELBO is a weighted version of this loss:
\begin{equation} \label{eq:S10}
    -\mathrm{ELBO}_\theta(\mathbf{a}, s) = T\,\mathbb{E}_{t,\varepsilon}\big[\lambda_t\, \ell_\theta(\mathbf{a}, s; t, \varepsilon)\big] + C
\end{equation}
where the weights $\lambda_t$ are fixed by the noise schedule and $C$ is independent of state $s$.
Since the ELBO is a tractable lower bound on $\log \pi_\theta(\mathbf{a} \mid s)$, we use its gradient as our estimate:
\begin{equation} \label{eq:S11}
    g_\theta(\mathbf{a}, s) \approx -T\,\nabla_s \mathbb{E}_{t, \varepsilon}\big[\lambda_t\, \ell_\theta(\mathbf{a}, s; t, \varepsilon)\big]
\end{equation}
In practice we evaluate the expectation over $t$ in \eq{S11} by importance sampling: $t$ is drawn from a proposal distribution $q(t)$ that favors low-noise steps (where the action distribution is more accurate), and each draw is weighted by $1/(T\,q(t))$.
This importance sampling approach leaves \eq{S11} unbiased while lowering its variance.

Once $g_\theta(\mathbf{a}, s)$ is estimated as shown above, we use it to compute the Fisher information matrix following \eq{S7}, where the expectation over actions is replaced by an average over $N_a$ actions sampled from the current policy, $\mathbf{a} \sim \pi_\theta(\cdot \mid s)$.
We require $N_a \geq r$ so that the estimated $F(s)$ can reach rank $r$.
Because \eq{S11} is evaluated with a finite number of $(t, \varepsilon)$ samples, the estimate of $g_\theta$ is noisy, and the outer product in \eq{S7} adds the covariance of this noise to $F(s)$.
This bias is positive semi-definite and shrinks as more samples are used.
It vanishes along any direction in which the denoising loss is insensitive to the state for every $(t, \varepsilon)$, so minimizing the estimated eigenvalues enforces a stronger version of our desired null space condition.
The error of our ELBO approximation is the gradient of the bound's gap, which vanishes when the learned reverse process exactly inverts the forward noise process.

Each of the $N_a$ sampled actions uses $N_\varepsilon$ draws of $(t, \varepsilon)$, and each draw requires one forward and one backward pass through the denoiser, so estimating $F(s)$ costs $N_a N_\varepsilon$ passes through the diffusion model per state.
The actions themselves are sampled via several steps of the reverse process without tracking gradients; we never need to differentiate through the sampling process when computing \eq{S11}.

\section{InneR Algorithm} \label{sec:algorithm}

Sections~\ref{sec:determ} and \ref{sec:stoch} theoretically outline how regularizing the state space inner product can make a learned policy $\pi_\theta$ robust to local perturbations $\Delta$.
Equations~(\ref{eq:J8}) and (\ref{eq:J9}) provide the essential steps: averaging the inner product over individual states, and then minimizing the eigenvalues which correspond to the shared null space.
However, thus far we have only discussed applying these steps across the entire dataset $\mathcal{D}$.
This naive implementation would encourage the policy to learn a consistent null space across all states --- i.e., encouraging every state to be robust to the same perturbations.
Global consistency makes sense for tasks where the robot must always attend to the same state elements (for example, picking up a block).
But, global regularization is undesirable when the relevant state information changes over the course of the task (for example, picking up a block and then stacking it on a target).

Our inner product regularization algorithm (\textbf{InneR}) addresses this challenge by first clustering the dataset into local regions $\mathcal{B}$ before shaping each region's average inner product.
In Section~\ref{sec:A1} we obtain local regions using inner product alignment.
For simplicity of explanation we focus on the inner product for deterministic policies, but the same steps apply to $F(s)$ with $m$ replaced by $r$ (Section~\ref{sec:S1}).
Next, in Section~\ref{sec:A2} we present the resulting InneR algorithm block.
We highlight that InneR can be appended to existing imitation learning architectures, and we provide guidance for parameters which the designer may need to tune.

\subsection{Clustering Based on Inner Product Alignment} \label{sec:A1}

To motivate localized regularization we start with an example.
Consider a table-top setting with two blocks and one cup; the system state could be broken into components for the robot and each object, i.e. $s = [s_\mathrm{robot}, s_\mathrm{block1}, s_\mathrm{block2}, s_\mathrm{cup}]$.
Imagine that the robot is learning to ``\textit{stack block1 on top of block2}.''
Throughout the entire motion the robot's actions should remain invariant to the cup's state.
At the start of the task the robot must focus on $s_\mathrm{block1}$, and modifications to $s_\mathrm{block2}$ can be ignored.
But once the robot grasps and lifts block1, then perturbations to $s_\mathrm{block2}$ ought to alter the robot's actions $a$.
Hence, whether a change in $s_\mathrm{block2}$ is relevant depends on the current context.

We refine our key insight in order to address this challenge.
Within Section~\ref{sec:D2} we argued that if a perturbation $\Delta$ does not impact the policy output at state $s$, that same perturbation should not impact the policy output at states near $s$.
Now we reverse that logic and define two states as nearby if they are both robust to similar perturbations.
Under this definition, nearby states need not be close in $\mathcal{S}$; rather, they are states at which the policy attends to similar state directions.
We propose to cluster states based on the alignment between their inner product matrices $A(s)$.
Groups of states that have the most aligned inner products are treated as \textit{regions}, and we locally apply our regularization to these regions to further increase their shared null space alignment.

\p{Inner Product Alignment}
\textit{Alignment} refers to the relative spaces spanned by matrices, and not the magnitude of those matrices.
To determine whether two arbitrary inner product matrices are aligned, we first compute the thin singular value decomposition of their corresponding Jacobians:
\begin{equation} \label{eq:A1}
    J(s_i) = U_i \Sigma_i V_i^T, \quad J \in \mathbb{R}^{m \times n}
\end{equation}
Here the columns of $V_i \in \mathbb{R}^{n \times m}$ are the right singular vectors of $J(s_i)$ associated with its $m$ nonzero singular values.
By construction the columns of $V_1$ and $V_2$ span the row spaces of $A(s_1)$ and $A(s_2)$, and we can use the Jacobian to obtain $V_i$ without forming the inner product matrix.

We next calculate the chordal Grassmann distance between the two row spaces \cite{conway1996packing}.
This distance depends only on the principal angles between two subspaces, independent of the specific basis vectors used to represent them.
Up to a constant factor, the chordal Grassmann distance is:
\begin{equation} \label{eq:A2}
    d_\mathrm{align}(A_1, A_2) = \| V_1V_1^T - V_2V_2^T\|_F
\end{equation}
where $\| \cdot \|_F$ is the Frobenius norm and $V_iV_i^T$ defines an orthogonal projection matrix for the row space.
Combining Equations~(\ref{eq:A1}) and (\ref{eq:A2}) quantifies the alignment between the row space of the inner product at different states.

\p{Clustering and Regularization}
Equipped with our ability to measure inner product alignment, we can now cluster states that have the most similar null spaces.
For each state $s_i$ in dataset $\mathcal{D}$ we compute $J(s_i)$.
We then apply Equations~(\ref{eq:A1}) and (\ref{eq:A2}) to get the pairwise distances with every other inner product matrix (resulting in a total of $K\times K$ pairwise distances).
Multiple clustering strategies are viable at this point, and our InneR algorithm is not tied to any specific approach.
However, because our experiments leverage $k$-nearest neighbors, we will focus on that method in what follows.
From the pairwise distances we identify the $k$ states with the lowest distance from $s_i$ (including $s_i$ itself, at distance zero), i.e., the $k$ inner products that are best aligned with $A(s_i)$.
This defines a local region $\mathcal{B}_i = \{s_\mathrm{near}^1, s_\mathrm{near}^2, \ldots s_\mathrm{near}^k\}$.
When performed in parallel across the dataset, we obtain $K$ regions $\mathcal{B}_1, \ldots, \mathcal{B}_K$, where the $i$-th region consists of state $s_i$ and the $k-1$ other states with the best aligned inner products.
In practice we form these regions within each training batch rather than across the full dataset: the pairwise distances are computed between the states of the batch, and the regions are recomputed at every gradient step as the policy changes (Section~\ref{sec:A2}).

Each region contains the inner products that are most closely aligned with $A(s_i)$. 
Our local regularization approach seeks to augment their alignment by applying the active subspace learning from Section~\ref{sec:D2}.
Building on \eq{J8}, we compute the average $A$ across every local region $\mathcal{B}_i$:
\begin{equation} \label{eq:A3}
    \bar{A}_i = \frac{1}{k} \sum_{s \in \mathcal{B}_i} A(s)
\end{equation}
We then apply \eq{J9} and extract the sorted eigenvalues for each of these averaged inner products:
\begin{equation} \label{eq:A4}
    \operatorname{eig}(\bar{A}_i) = (\lambda_1^i, \ldots, \lambda_m^i, \lambda_{m+1}^i, \ldots, \lambda_n^i)
\end{equation}
The shared null space for the $i$-th region corresponds to the last $n-m$ elements $\lambda_{m+1}^i, \ldots, \lambda_n^i$.
Accordingly, to shape a policy so that it is \textit{locally} robust to similar perturbations, our regularization loss penalizes the fraction of each region's inner product that lies in the shared null space:
\begin{equation} \label{eq:A5}
    \mathcal{L}_\mathrm{InneR}(\theta) = \frac{1}{K} \sum_{i = 1}^K \frac{\sum_{j = m+1}^n \lambda^i_j}{\operatorname{sg}\big(\sum_{j = 1}^n \lambda^i_j\big)}
\end{equation}
where $\operatorname{sg}(\cdot)$ is the stop-gradient operator which treats its argument as a constant when differentiating.

We note that eigenvalues $\lambda$ are a function of the policy weights $\theta$.
Each term in \eq{A5} lies between $0$ and $1$ regardless of the scale of the policy Jacobian.
This normalization matters: the raw eigenvalues $\lambda^i_{m+1}, \ldots, \lambda^i_n$ can be driven to zero simply by shrinking $J(s)$ everywhere, which is the behavior of existing norm penalties in Section~\ref{sec:D2} that we seek to avoid.
Dividing by the total removes this incentive, and detaching the total prevents the loss from being lowered by inflating the leading eigenvalues instead of reducing the smallest ones.
Because the denominator is the trace of $\bar{A}_i$, the normalization adds no computational cost, and it also makes the weight $\gamma$ (described below) more comparable across tasks and policy classes.
To combine our regularizer with existing imitation learning algorithms we simply augment the given loss $\mathcal{L}$.
The updated loss adds our shared null space penalty:
\begin{equation} \label{eq:A6}
    \mathcal{L}(\theta) \gets \mathcal{L}(\theta) + \gamma \cdot \mathcal{L}_\mathrm{InneR}(\theta)
\end{equation}
where $\gamma \geq 0$ is a hyperparameter that determines the weight of the regularizer.
We discuss this (and the effects of other parameter selection) in Section~\ref{sec:A2}.

\p{Extension to Stochastic Policies}
Within the steps above we focused on deterministic policies with inner product $A(s)$.
However, the same alignment and clustering approaches can be applied to stochastic policies by replacing $A(s)$ with $F(s)$ and $m$ with $r$ (Section~\ref{sec:S1}).
Matrix $V_i$ then contains the $r$ leading eigenvectors of $F(s_i)$, which we obtain as the leading right singular vectors of the matrix of sampled scores $g_\theta(\mathbf{a}, s_i)$ without forming $F(s_i)$, and \eq{A5} penalizes the $n-r$ smallest eigenvalues of each $\bar{F}_i$ relative to its detached trace.

\subsection{Algorithm and Implementation} \label{sec:A2}

\begin{algorithm}[t]
    \caption{InneR for Deterministic Policies}
    \label{alg:1}
    \begin{algorithmic}[1]
    \Require Dataset $\mathcal{D}$, policy $\pi_\theta$, action space inner product $M$, neighbors $k$, regularization weight $\gamma$
    \While{not converged}
        \State Sample batch $\{(s_i, a_i)\}_{i=1}^{B} \sim \mathcal{D}$
        \State Compute base loss $\mathcal{L}$ across batch
        \For{each state $s_i$ in batch}
            \State $J_i \gets \partial \pi_\theta / \partial s \,|_{s_i}$ \Comment{state Jacobian}
            \State $A_i \gets J_i^T M J_i$  \Comment{get inner products}
        \EndFor
        \State \textbf{with} $\text{no\_grad}$ \textbf{do} \Comment{cluster by alignment}
        \State \hspace{\algorithmicindent} $V_i \gets$ right singular vectors of $J_i$
        \State \hspace{\algorithmicindent} $d_{ij} \gets \| V_i V_i^T - V_j V_j^T \|_F$ \Comment{\small Grassmann distance}
        \State \hspace{\algorithmicindent} $\mathcal{N}_i \gets$ indices of the $k$ smallest $d_{ij}$ (including $i$)
        \State \textbf{end with}
        \For{each state $s_i$ in batch}
            \State $\bar{A}_i \gets \frac{1}{k} \sum_{j \in \mathcal{N}_i} A_j$ \Comment{average inner product}
            \State $\lambda^i_1 \geq \cdots \geq \lambda^i_n \gets$ eigenvalues of $\bar{A}_i$
            \State $\rho_i \gets \sum_{j = m+1}^{n} \lambda^i_j \,\big/\, \operatorname{sg}\big(\sum_{j = 1}^{n} \lambda^i_j\big)$
        \EndFor
        \State $\mathcal{L}_\mathrm{InneR} \gets \frac{1}{B} \sum_{i=1}^{B} \rho_i$ \Comment{\small penalize the shared null space}
        \State $\mathcal{L} \gets \mathcal{L} + \gamma \cdot \mathcal{L}_\mathrm{InneR}$ \Comment{augment base loss}
        \State Update policy parameters $\theta$ to minimize loss $\mathcal{L}$
    \EndWhile
    \end{algorithmic}
\end{algorithm}

An outline of InneR is provided in Algorithm~\ref{alg:1}.
After computing the base loss for the given imitation learning method, InneR adds the following steps: 
calculating the Jacobians and inner products for each state in the batch (\textit{lines 4-6}), clustering the states based on inner product alignment (\textit{lines 7-10}), and then computing the fraction of each region's inner product that lies in a shared null space (\textit{lines 12-16}).

\p{Hyperparameter Tuning}
InneR introduces three parameters the designer must set: $M$, $k$, and $\gamma$ (and, for stochastic policies, the rank $r$ from Section~\ref{sec:S1}).
Matrix $M$ should default to identity.
If the designer wants to make the policy more sensitive to mistakes in a specific action dimension (e.g., opening or closing a gripper), then the designer should increase the relative value of that diagonal element of $M$.
Scalar $k \geq 2$ is the number of nearest neighbors for clustering.
As $k$ increases to the batch size $B$, InneR encourages a globally consistent null space; conversely, as $k \rightarrow 2$ the regions become increasingly local.
In our experiments we set $k=8$.
Finally, $\gamma \geq 0$ is the weight of the added InneR loss.
Because the loss is normalized in \eq{A5}, designers should not need to significantly change $\gamma$ when working with different state dimensions.
We found $0.1 \leq \gamma \leq 0.5$ to be effective in single-integrator simulations and hardware experiments, and $0.01 \leq \gamma \leq 0.05$ to be effective in drone-landing simulations.
Larger values let the regularizer dominate early in training --- before the policy has learned which state directions matter --- and can collapse every state onto a single shared null space.

\p{Computational Cost}
InneR increases the time required for training an imitation learning policy.
In our testing with a 16-dimensional state in Section~\ref{sec:sims}, InneR increased the training time by 6.5x on a multilayer perceptron (MLP) and 5x on a diffusion policy.
The added cost has three sources.
First, the inner product at every state of the batch requires additional passes through the policy with the computation graph retained: $m$ backward passes to form $J(s)$ for the MLP, and $N_a N_\varepsilon$ forward and backward passes of the denoiser (plus sampling the $N_a$ actions) to estimate the scores for the diffusion policy.
Second, because $\mathcal{L}_\mathrm{InneR}$ depends on these derivatives, its gradient must be propagated back through them, i.e., a second-order derivative of the policy; this backward pass is the single most expensive step for both policy classes (roughly a third of each gradient step's complexity).
Finally, clustering and regularizing the batch require $B$ singular value decompositions, $B^2$ subspace distances, and $B$ eigendecompositions of $n \times n$ matrices.
For a deterministic MLP policy, the additional derivatives and linear algebra computations each account for roughly half of the added time; with a diffusion policy the score estimate accounts for roughly three quarters.

\section{Single-Integrator Simulations} \label{sec:sims}

\begin{figure*}[t]
    \centering
    \includegraphics[width=0.9\linewidth]{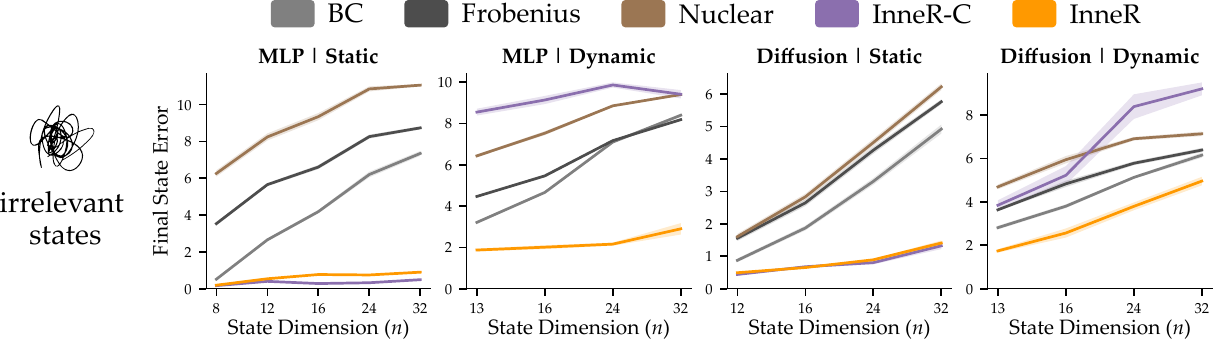}
    \caption{Single-integrator simulation from Section~\ref{sec:sim1}. Plots show the error between the robot and the goal as state dimension $n$ increases (lower is better). The far left of each plot corresponds to the minimum state dimension for that setting, and each added state element is irrelevant noise that is sampled uniformly at random in both the demonstrations and the test states. Ideally the robot is robust to these variations in the irrelevant state dimensions, and so the final state error should be constant in $n$. \textbf{InneR} is the only method whose error stays nearly flat in every setting besides Diffusion--Dynamic, where it is consistently lower than the alternatives while maintaining a similar positive slope. \textbf{InneR-C} matches it in \textit{static}, where one global null space is correct, but fails in \textit{dynamic}. Shaded regions show standard error across $10$ seeds.}
    \label{fig:irrelevant}
    \vspace{-0.5em}
\end{figure*}

\begin{figure*}[t]
    \centering
    \includegraphics[width=1\linewidth]{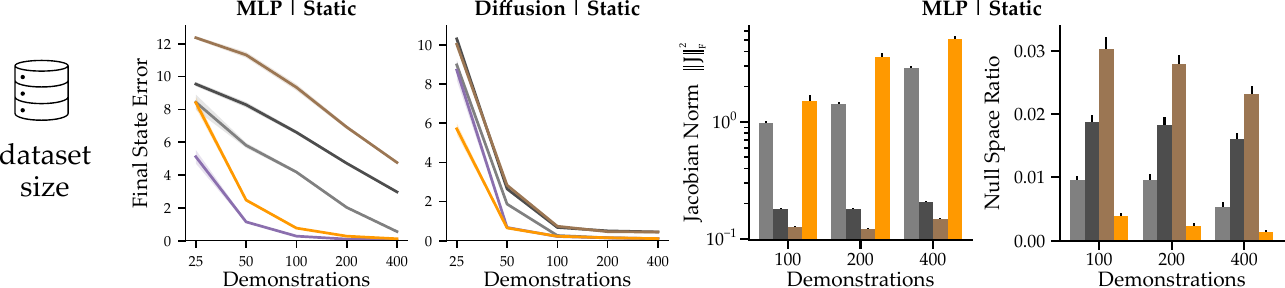}
    \caption{Single integrator simulation from Section~\ref{sec:sim2} on the \textit{static} task with state dimension $n=16$. (Left) Final state error as the number of demonstrations increases for the MLP and Diffusion policies. Here \textbf{InneR} and \textbf{InneR-C} are both effective since the same state entries always contain relevant or irrelevant information. (Right) What each method regularizes, measured for the final MLP policy at $200$ training states: the squared Frobenius norm of the Jacobian (log scale) and the shared null space ratio of \eq{A5} (\textbf{InneR-C} matches \textbf{InneR}, and is not shown for brevity). Standard regularizers like \textbf{Frobenius} and \textbf{Nuclear} are designed to decrease the magnitude of each Jacobian, and by flattening these individual Jacobians they actually make the perturbation responses inconsistent across the dataset: their null space ratio rises above the \textbf{BC} value. By contrast, \textbf{InneR} leaves $\|J(s)\|^2_F$ for each individual state Jacobian at or above its \textbf{BC} value, and locally aligns the Jacobians so that they have a consistent null space. Colors follow \fig{irrelevant}.}
    \label{fig:data}
    \vspace{-1em}
\end{figure*}

\begin{figure}[t]
    \centering
    \includegraphics[width=1\columnwidth]{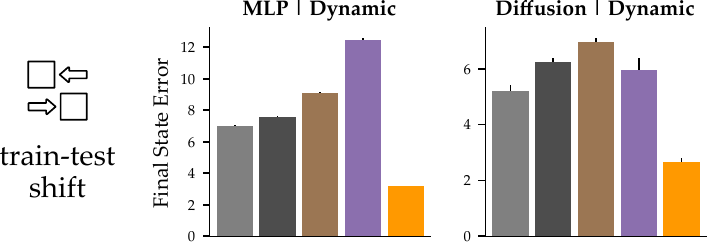}
    \caption{Single-integrator simulation from Section~\ref{sec:sim3} highlighting the \textit{dynamic} task. At training time the irrelevant state dimensions are correlated with the active goal ($\rho = 0.9$), but at test time this correlation is removed, so a policy that reads the goal from the irrelevant entries fails (i.e., causal confusion \cite{dehaan2019causal}). \textbf{InneR} is the only method that improves on \textbf{BC}.}
    \label{fig:shift}
    \vspace{-1em}
\end{figure}

\begin{figure}[t]
    \centering
    \includegraphics[width=1\columnwidth]{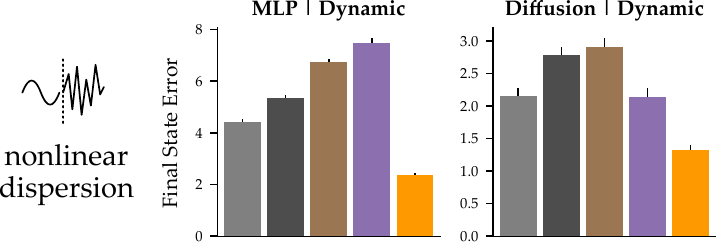}
    \caption{Single-integrator simulation from Section~\ref{sec:sim4} highlighting the \textit{dynamic} task. An invertible nonlinear transformation is applied to the state before it is input to the policy, so that the task-relevant information is dispersed across every entry the policy observes and the dynamics are nonlinear in those coordinates. \textbf{InneR} lowers the error of \textbf{BC} despite this nonlinear dispersion.}
    \label{fig:scramble}
    \vspace{-1.5em}
\end{figure}

We begin by conducting simulation experiments in a point-mass environment to investigate explore InneR's behavior in a controlled setting.
All code for reproducing these simulations is available on our project site: \url{https://collab.me.vt.edu/inner}.

\p{Setup}
We study a goal reaching task where the robot is modeled as a single-integrator with linear dynamics.
The state dimension is $n=16$ ($n$ is varied in some simulations) and the action dimension is $m=4$.
The state $s$ concatenates the position of the robot ($m$ entries), the position of the goal ($m$ entries), and $n - 2m$ irrelevant noise entries.
The action $a$ is the robot's displacement: at each timestep the robot's position is incremented by $a$ while the goal and noise remain fixed.
During our experiments the initial state is uniformly randomly sampled from $[-10, 10]^n$ for both data collection and testing.
In each demonstration the expert shows a trajectory of 40 state-action pairs that move directly from the start to the goal at a speed of at most one unit per timestep (the total number of state-action pairs in $\mathcal{D}$ is equal to 40 times the number of demonstrations).
At test time the robot is similarly given 40 timesteps to reach the goal using its learned policy.
We report \textit{final state error}, the average distance between the robot's final state and the goal across 1000 random test starts.

\p{Static and Dynamic Goals}
There are two versions of the point-mass environment.
In \textit{static} the goal position always occupies the same $m$ entries of the state.
In \textit{dynamic} there are two goals in the environment, and the correct goal for the current trial is signaled by the sign of the final state element: if this switch is positive the robot should reach the first goal, and if negative the robot reaches the second.
The second goal occupies the $m$ entries after the first, so that $n - 3m - 1$ entries remain noise.
\textit{Static} is an example of a task where the policy's response to state perturbations can be globally consistent (i.e., the same state elements are always relevant or irrelevant).
By contrast, in \textit{dynamic} the relevant elements of the state change in different regions, motivating a local regularization approach.

\p{MLP and Diffusion Policies}
For each experiment we test a deterministic policy (\textit{MLP}) and a stochastic policy (\textit{Diffusion}).
MLP has two hidden layers of width $256$.
Diffusion is a denoising diffusion model with $T = 100$ steps \cite{ho2020denoising, chi2025diffusion}; its noise prediction network has three hidden layers of width $256$ and outputs a single action per state (i.e., $H = 1$ and $n_0 = 1$ in Section~\ref{sec:S2}).
We sample actions with $10$ DDIM steps at test time, and estimate the Fisher information in \eq{S7} with $N_a = 8$ sampled actions and $N_\varepsilon = 4$ draws of $(t, \varepsilon)$ per action.
Both MLP and Diffusion policies are trained with Adam \cite{kingma2015adam} and a batch size of $B = 200$: the MLP for $200$ epochs at a learning rate of $10^{-3}$, and the diffusion policy for $1000$ epochs at $3 \cdot 10^{-3}$.
These settings are held constant across methods and experiments.

\p{Baselines}
We test InneR with $M=I$, $k=8$, and $\gamma=0.2$ for MLP; for Diffusion we increase $\gamma=0.3$ and conservatively set $r=2m=8$.
\textbf{InneR} is compared to four baselines:
standard behavior cloning (\textbf{BC}), \textbf{Frobenius} norm regularization \cite{hoffman2019robust}, \textbf{Nuclear} norm regularization \cite{scarvelis2024nuclear}, and \textbf{InneR-C}.
All algorithms are given the exact same training data, model architecture, training hyperparameters, and start states at test time.
For \textbf{Frobenius} and \textbf{Nuclear} the loss is the \textbf{BC} loss plus $0.01$ times the average of the squared Frobenius norm or nuclear norm of each individual Jacobian.
For the diffusion policy we use analogous quantities, the trace of $F(s)$ (weighted by $10^{-4}$) and the sum of the square roots of its eigenvalues (weighted by $10^{-3}$).
We selected these weights by sweeping over several orders of magnitude; larger weights monotonically increased the final state error.
\textbf{InneR-C} is an ablation of our method without clustering.
In practice, \textbf{InneR-C} is equivalent to Algorithm~\ref{alg:1} with $k=B$ so that each local region is the entire batch.
We repeat every experiment over $10$ random seeds; the figures report the mean and standard error across these seeds, with the colors of \fig{irrelevant} used throughout.

\subsection{How Does InneR Scale with Irrelevant State Dimension?} \label{sec:sim1}

In \fig{irrelevant} we plot final state error as a function of the number of state dimensions, where every added state coordinate provides information irrelevant to the task.
For \textit{static} there are $n-2m$ irrelevant state dimensions, and for \textit{dynamic} there are $n-2m-1$ irrelevant dimensions (since the goal not selected is irrelevant).
We fix the number of demonstrations at a value where \textbf{BC} does not solve the $n = 16$ task ($100$ and $400$ demonstrations for the MLP on \textit{static} and \textit{dynamic}, $50$ and $200$ for the diffusion policy).
The error of \textbf{BC} grows steadily with every irrelevant dimension (for the MLP on \textit{static}, from $0.5$ at $n = 8$ to $7.4$ at $n = 32$), and the norm baselines are worse than \textbf{BC} at every $n$.
\textbf{InneR} stays nearly flat (growing only from $0.2$ to $0.9$) and is the only method below \textbf{BC} in both \textit{dynamic} settings.
As expected, we find that \textbf{InneR-C} is effective in \textit{static} where the same state dimensions are always relevant, but falls short in \textit{dynamic} where the relevant perturbations change across the state space.

\subsection{How Does InneR Scale with Dataset Size?} \label{sec:sim2}

In \fig{data} we focus on the \textit{static} task, where a single shared null space is correct for every state, and plot final state error as a function of the number of demonstrations (on \textit{dynamic} the methods order as in \fig{irrelevant}).
\textbf{InneR} and \textbf{InneR-C} bring the error below one unit with $100$ demonstrations for the MLP and $50$ for the diffusion policy, whereas \textbf{BC} needs $400$ and $100$.
To better understand these results, we also plot what each method is actually regularizing.
We see that the norm baselines shrink the squared Frobenius norm of the Jacobian far below its value under standard \textbf{BC}, whereas \textbf{InneR} leaves this norm at or above the \textbf{BC} value (the trace of $F$ for the diffusion policy behaves the same way, though it is not shown for brevity).
On the other hand, \textbf{InneR} minimizes the fraction of the averaged inner product that lies in the shared null space, which the norm baselines actually increase.
These results indicate that \textbf{InneR} effectively retains the expressiveness of the given policy while regulating its shared null space.

\subsection{Is InneR Robust to Shifted Training and Test Distributions?} \label{sec:sim3}

Next, we collect training data where the irrelevant state dimensions are correlated with the intended goal.
Each irrelevant entry of a demonstration start state is a convex combination of its own uniform draw and one coordinate of the active goal, mixed so that their Pearson correlation is $\rho = 0.9$; at test time the initial state is uniformly sampled without this spurious correlation.
The correlation allows the policy to estimate the goal from the irrelevant entries, but a policy that relies on this shortcut fails once it disappears at test time.
Our results for \textit{dynamic} with $400$ demonstrations for the MLP and $300$ for the diffusion policy are shown in \fig{shift}.
\textbf{InneR} roughly halves the error of \textbf{BC} for both policies ($3.2$ against $7.0$ for the MLP, $2.7$ against $5.2$ for Diffusion), while every other regularizer is worse than \textbf{BC}.
\textbf{InneR-C} fails most severely for the MLP, since the one global null space it enforces cannot cope with the goal switch.

\subsection{What if Information is Dispersed Throughout the State?} \label{sec:sim4}

So far our experiments separate the state dimensions into relevant and irrelevant entries, meaning that the shared null space is axis-aligned.
Here we apply a nonlinear mapping to the underlying state $s$ so that the relevant information is scrambled throughout the states $z$ that the policy inputs.
We apply a per-entry tanh bend between two affine layers, $z = A_2 \phi(A_1s + b_1) + b_2$, where $A_1$ is a random matrix with condition number at most $3$, $A_2$ is a random rotation, $b_1$ and $b_2$ are random offsets, and the $i$-th coordinate of the nonlinear transformation is given by $\phi_i(x) = x + c_i \tau \tanh(x / \tau)$ with $\tau = 5$, $c_i \in [-0.5, 0.5]$, and argument $x$ is the $i$-th coordinate of the vector-valued input of $\phi$.
This mapping is invertible (so that no data is lost) and is drawn once per random seed; every entry of $z$ now mixes relevant and irrelevant entries of $s$.
In addition, the system dynamics are now nonlinear in the coordinates the policy observes.
Our results for \textit{dynamic} with $800$ demonstrations for the MLP and $600$ for the diffusion policy are shown in \fig{scramble}.
\textbf{InneR} lowers the error of \textbf{BC} from $4.4$ to $2.4$ for the MLP and from $2.2$ to $1.3$ for Diffusion, whereas the norm baselines again increase it.


\section{Drone Landing Simulations}
\label{sec:drone_sim}

\begin{figure}[t]
    \centering
    \includegraphics[width=1\linewidth]{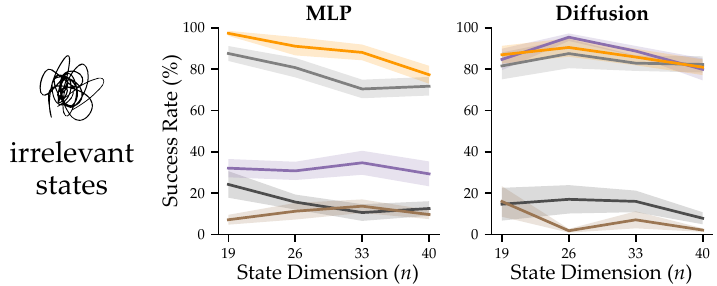}
    \caption{Drone landing simulation from Section~\ref{sec:drone_sim1}. Plots show the landing success rate as state dimension $n$ increases (higher is better). The state dimension sweep starts at $n=19$, and each additional $7$ state coordinates come from a distractor drone in both the demonstrations and the test states. \textbf{InneR} largely outperforms \textbf{BC} at all tested dimensions for MLP, and is close to \textbf{BC} for Diffusion. \textbf{InneR-C} performs poorly for MLP but is competitive for Diffusion.}
    \label{fig:drone_irrelevant}
    \vspace{-0.5em}
\end{figure}

\begin{figure*}[t]
    \centering
    \includegraphics[width=1\linewidth]{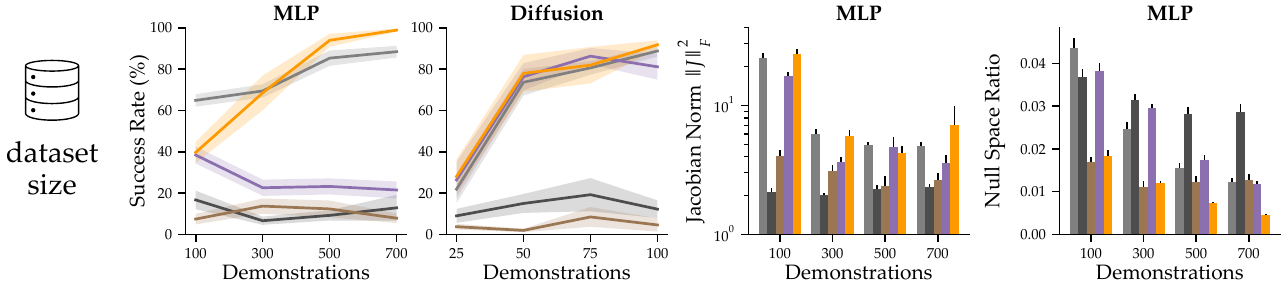}
    \caption{Drone landing simulation from Section~\ref{sec:drone_sim2}. Plots show the landing success rate as number of demonstrations increases (higher is better) with state dimension $n=22$.
    (Left) Success rate as the number of demonstrations increases for the MLP and Diffusion policies. 
    \textbf{InneR} is effective for both MLP and Diffusion policies, and \textbf{InneR-C} is only effective for the Diffusion policy.
    (Right) What each method regularizes (measured for the final MLP policy at $200$ training states), i.e. the squared Frobenius norm of the Jacobian (log scale) and the shared null space ratio of \eq{A5}.
    Both \textbf{Frobenius} and \textbf{nuclear} norm baselines reduce the Jacobian norm. \textbf{Frobenius} has a higher null space ratio than BC at 300, 500, and 700 demonstrations, whereas \textbf{Nuclear} is below BC except at 700 demonstrations. \textbf{InneR} consistently reduces the null-space ratio without uniformly reducing the Jacobian norm. Colors follow \fig{irrelevant}.}
    \label{fig:drone_data}
    \vspace{-1em}
\end{figure*}

\begin{figure}[t]
    \centering
    \includegraphics[width=1\columnwidth]{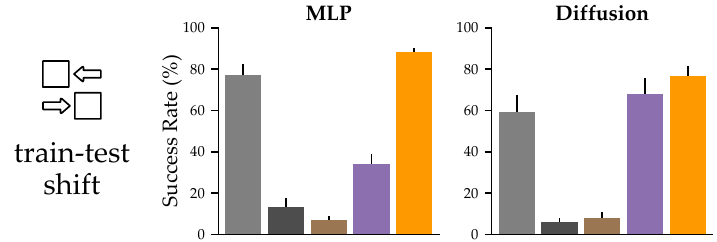}
    \caption{Drone landing simulation from Section~\ref{sec:drone_sim3}. In training, irrelevant state dimensions are correlated with the active goal ($\rho = 0.9$), but at test time this correlation is removed, so a policy that reads the goal from the irrelevant entries fails (i.e., causal confusion \cite{dehaan2019causal}). \textbf{InneR} improves upon \textbf{BC} for both MLP and Diffusion, while \textbf{InneR-C} only improves upon \textbf{BC} for Diffusion.}
    \label{fig:drone_shift}
    \vspace{-1em}
\end{figure}

\begin{figure}[t]
    \centering
    \includegraphics[width=1\columnwidth]{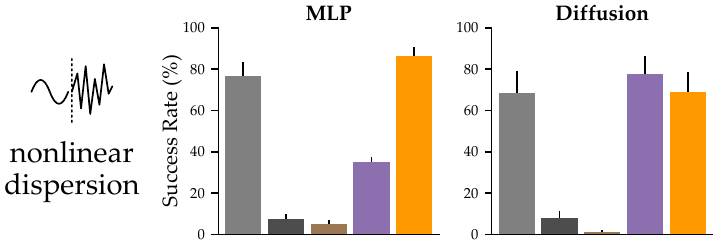}
    \caption{Drone landing simulation from Section~\ref{sec:drone_sim4}. An invertible nonlinear transformation is applied to the state before it is input to the policy, so that the task-relevant information is dispersed across every entry the policy observes and the dynamics are nonlinear in those coordinates. \textbf{InneR} improves upon \textbf{BC} for MLP and \textbf{InneR-C} improves upon \textbf{BC}  for Diffusion despite this nonlinear dispersion.}
    \label{fig:drone_scramble}
    \vspace{-1.5em}
\end{figure}

Having investigated the performance of InneR in the controlled single-integrator environment, we next consider more realistic simulations involving a drone-landing task with nonlinear dynamics.

\p{Setup}
In the drone-landing task, a Crazyflie 2.0 \cite{giernacki2017crazyflie} quadrotor simulated in the \texttt{gym-pybullet-drones} \cite{panerati2021learning} environment must fly toward a specified landing point, descend, and come to rest on the ground.
The action $a_t\in\mathbb{R}^4$ specifies the desired rotation rate for the drone's four motors. The state dimension is $n=22$ ($n$ is varied in some simulations), in which the drone's state $s_t\in\mathbb{R}^{19}$ is
\begin{equation*}
    s_t=\begin{bmatrix}
        p_t^\top & q_t^\top & \theta_t^\top & v_t^\top & \omega_t^\top & p_g^\top
    \end{bmatrix}^\top,
\end{equation*}
consisting of the drone's position ($p_t\in\mathbb{R}^3$),  orientation specified in both  quaternion ($q_t\in\mathbb{R}^4$) and Euler angle ($\theta_t\in\mathbb{R}^3$) formats, linear velocity ($v_t\in\mathbb{R}^3$), angular velocity ($\omega_t\in\mathbb{R}^3$) and goal position ($p_g\in\mathbb{R}^3$). The remaining $n-19$ dimensions are task-irrelevant and will be introduced in greater detail below.

For both training and testing, the initial horizontal position is sampled uniformly from ${[-0.1\si{\meter},0.1\si{\meter}]\times[-0.1\si{\meter},0.1\si{\meter}]}$, with altitude fixed at $0.7\si{\meter}$, zero initial velocity, and level orientation. The static target $p_g$ is sampled uniformly with ${p_{g,x}\in[0.25\si{\meter},0.45\si{\meter}]}$ and
${p_{g,y}\in[-0.25\si{\meter},0.25\si{\meter}]}$. Its height is fixed at ${p_{g,z}=0.0125\si{\meter}}$, corresponding to the drone's nominal center height at rest.

An expert controller first guides the drone toward the target, then descends and switches off its motors after touchdown. It commands a desired velocity with magnitude at most $0.25\si{\meter}/\si{\second}$ and produces
motor commands at $48\ \text{Hz}$, while the simulator advances at $240\ \text{Hz}$. Each demonstration contains at most $576$ state-action pairs, corresponding to $12$ seconds of demonstration, including up to one second of recording after touchdown.

At test time, each policy receives at most $960$ control steps ($20$ \si{\second}), to land. Success requires ground contact together with horizontal error at most $0.05\si{\meter}$, vertical error at most $0.06\si{\meter}$, speed at most $0.15\si{\meter}/\si{\second}$, and roll-pitch magnitude at most $\pi/9\ \text{rad}$, maintained for ten consecutive control steps.

\p{MLP and Diffusion Policies}
For each experiment, we test a deterministic \textit{MLP} policy and a stochastic \textit{Diffusion} policy.
\textit{MLP} has two hidden layers of width $128$ with ReLU activations.
\textit{Diffusion} is a denoising diffusion model with $T=100$ steps \cite{ho2020denoising, chi2025diffusion}; its noise prediction network has three hidden layers of width $128$ with ReLU activations.
Both policies map the current observation to a single four-dimensional motor command, with $H=1$ and $n_0=1$ for the diffusion policy in Section~\ref{sec:S2}.
We sample diffusion actions using $10$ DDIM \cite{song2020denoising} steps at test time, and estimate the Fisher information in \eqref{eq:S7} with $N_a=8$ sampled actions using $3$ DDIM steps and $N_\varepsilon=4$ draws of $(t,\varepsilon)$ per action.
Both policy architectures are trained with Adam \cite{kingma2015adam} using a batch size of $B=200$, and a learning rate of $10^{-3}$ for $100$ epochs.
Within each policy family and experimental condition, all compared methods use the same architecture, training demonstrations, training settings and test cases and are evaluated at the last epoch.

\p{Baselines}
We test \textbf{InneR} using $k=8$ and retained rank $r=m=4$ for MLP, $k=32$ and $r=2m=8$ for Diffusion, and $M=I$ and $\gamma=0.01$ for both policies.
\textbf{InneR} is compared to four baselines: unregularized behavior cloning (\textbf{BC}), \textbf{Frobenius} norm regularization \cite{hoffman2019robust}, \textbf{Nuclear} norm regularization \cite{scarvelis2024nuclear}, and \textbf{InneR-C}.
\textbf{Frobenius} uses $\gamma=0.01$ for MLP and $10^{-4}$ for Diffusion. \textbf{Nuclear} uses $\gamma=0.01$ for MLP and $10^{-3}$ for Diffusion.
As above, \textbf{InneR-C} is an ablation of our method without clustering (equivalent to setting $k = B$), keeping $M, \gamma, r$ unchanged.
We report the mean and standard error across 10 random seeds for each experiment.

\subsection{How does InneR Scale with Irrelevant State Dimension?}
\label{sec:drone_sim1}
In \fig{drone_irrelevant}, we plot landing success rate as a function of the observation dimension $n\in\{19,26,33,40\}$ while fixing the training set at $500$ demonstrations for MLP and $100$ for Diffusion.
Starting from the $19$-dimensional drone state and goal, we append the recorded poses of zero to three distractor drones.
Each pose contributes seven entries: three position coordinates and a four-dimensional quaternion.
These observations vary over time but do not affect the controlled drone's dynamics or landing objective.
We evaluate all methods on the same $250$ held-out test cases and report success rate (SR) using the aforementioned success criteria. 

\begin{figure*}[t]
    \centering
    \includegraphics[width=1.0\linewidth]{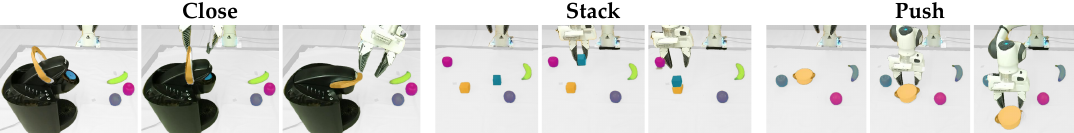}
    \caption{Our real-world tasks. Each task is shown at the start, middle, and end of a successful \textbf{InneR} rollout with the three distractors (apple, orange and banana). (Left) in \textit{close}, the robot pulls the handle down to latch the lid of the coffee maker. (Middle) In \textit{stack}, the robot picks up the black cube and places it on the tan cube (teal and orange segmentations, respectively). (Right) In \textit{push}, the robot pushes the pot into the target region. Tracked objects are tinted by the segmentation masks that our perception pipeline uses to estimate their poses, and the background is faded for clarity.}
    \label{fig:hw-tasks}
    \vspace{-1.0em}
\end{figure*}

For MLP, \textbf{InneR} has a higher average success rate than \textbf{BC} at every tested dimension, achieving $97.3\%$, $91.0\%$, $88.0\%$, and $77.2\%$ at $n=19$, $26$, $33$, and $40$, respectively.
For Diffusion, \textbf{InneR} improves on \textbf{BC} at $n=19$, $26$, and $33$.
At $n=40$, their mean success rates are similar: $80.9\%$ for \textbf{InneR} and $82.2\%$ for \textbf{BC}.
\textbf{InneR-C} has a low success rate for MLP but remains competitive for Diffusion, suggesting that the nonlinear drone dynamics require smaller clusters than were required in the linear single-integrator dynamics of Section \ref{sec:sims} when using deterministic policies. 
For both MLP and Diffusion policies, the \textbf{Frobenius} and \textbf{Nuclear} baselines attain a low average success rate.

\subsection{How Does InneR Scale with Dataset Size?}
\label{sec:drone_sim2}
In \fig{drone_data}, we plot landing success rate as a function of the number of training demonstrations: $\{100,300,500,700\}$ for MLP, and $\{25,50,75,100\}$ for Diffusion.
The policy input contains the $19$-dimensional drone state and goal together with three irrelevant entries sampled uniformly from $[-1,1]$ and held fixed within each episode, giving $n=22$.
Within each policy family, all methods receive the same training demonstrations and are evaluated on the same test cases.

For MLP policies, \textbf{InneR} improves on \textbf{BC} at larger numbers of demonstrations ($500$ and $700$), reaching $98.8\%$ success at $700$ demonstrations compared with $88.4\%$ for \textbf{BC}.
For Diffusion policies, \textbf{InneR} improves on \textbf{BC} at all tested demonstration dataset sizes.
\textbf{InneR-C} performs poorly for the MLP policy but remains competitive for the Diffusion policy. 
The \textbf{Frobenius} and \textbf{Nuclear} baselines have lower mean success rates than \textbf{BC} at every tested dataset size in both policy families.
To examine what the regularizers change, we also plot the squared Frobenius norm of the MLP policy Jacobian and the InneR null-space ratio.
The \textbf{Frobenius} and \textbf{Nuclear} norm baselines reduce the Jacobian norm below \textbf{BC} at every dataset size.
In contrast, \textbf{InneR} reduces the null space ratio relative to \textbf{BC} at all four dataset sizes.
As in Section~\ref{sec:sim2}, these results distinguish suppressing overall policy sensitivity from regularizing sensitivity in shared null space directions.

\subsection{Is InneR Robust to Shifted Training and Test Distributions?}
\label{sec:drone_sim3}

Next, we collect training data in which irrelevant observations are correlated with the landing target.
The $22$-dimensional input contains the $19$-dimensional drone state and goal together with a three-dimensional distractor goal that is held fixed within each episode.
During training, the horizontal distractor coordinates are generated to have correlation $\rho=0.9$ with the corresponding true-goal coordinates.
At test time, this correlation is removed while the true goal remains available in the observation.
We use $500$ demonstrations for MLP and $100$ for Diffusion, and evaluate each policy on the same test cases within its policy family.

\fig{drone_shift} plots the success rate under the shifted test distribution.
\textbf{InneR} achieves the highest mean success rate in both policy families under the shifted test distribution.
It outperforms \textbf{BC} ($88.4\%$ vs. $77.0\%$ for MLP and $76.6\%$ vs. $59.4\%$ for Diffusion).
\textbf{InneR-C} improves Diffusion success to $68.0\%$ but reduces MLP success to $34.3\%$; both \textbf{Frobenius} and \textbf{Nuclear} baselines perform worse than \textbf{BC} in both policy families.
These results support improved landing reliability for \textbf{InneR} under the tested shift.

\subsection{What if Information is Dispersed Throughout the State?}
\label{sec:drone_sim4}
Next, following Section~\ref{sec:sim4} we mix relevant and irrelevant information through an invertible nonlinear transformation of the $22$-dimensional observation from the dataset-size experiment.
We use $500$ training demonstrations for MLP policies and $100$ for Diffusion policies.
After training set normalization, the policy receives input $z=A_2\phi(A_1s+b_1)+b_2$, where we construct the nonlinearity elementwise: $\phi_i(x)=x+c_i\tau\tanh(x/\tau)$ with $A_1$ having condition number at most $3$.
The matrix $A_2$ is orthogonal and chosen independently in each seed; entries of $b_1$ and $b_2$ are sampled uniformly from $[-1,1]$.
We use $c_i\sim\mathcal{U}[-0.5,0.5]$ and $\tau=5$.
The mapping preserves the observation's information while dispersing it across coordinates.
One transformation is drawn per seed and held fixed across methods, training, and testing.
Regularization is computed with respect to the transformed observation $z$, and the drone dynamics, expert actions, and landing objective remain unchanged.

\fig{drone_scramble} reports the success rate over the dispersion, with $250$ test cases per run.
For MLP, \textbf{InneR} outperforms \textbf{BC} ($86.4\%$ vs. $76.8\%$).
\textbf{InneR-C} achieves $35.0\%$, while \textbf{Frobenius} and \textbf{Nuclear} baselines achieve $7.5\%$ and $5.2\%$, respectively.
For Diffusion, \textbf{InneR} and \textbf{BC} have similar mean success rates ($69.1\%$ and $68.4\%$), whereas \textbf{InneR-C} achieves the highest rate at $77.6\%$.
Both \textbf{Frobenius} and \textbf{Nuclear} norm baselines have near-zero success.
Thus, \textbf{InneR} retains high MLP success after information is dispersed, while \textbf{InneR-C} gives the largest mean Diffusion improvement in this setting.


\begin{figure*}[t]
    \centering
    \includegraphics[width=1.0\linewidth]{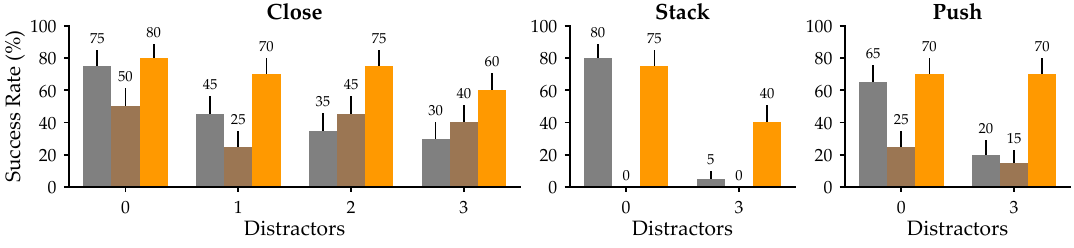}
    \caption{Success rates in our real-world experiments for \textbf{BC}, \textbf{Nuclear}, and \textbf{InneR}; the number above each bar is the success rate in percent. Colors follow \fig{irrelevant}.
    (Left) \textit{Close} with zero to three distractors. 
    (Middle, Right) \textit{Stack} and \textit{push} with zero and three distractors. 
    Error bars show standard error over $20$ trials. 
    Averaged across the three tasks, \textbf{InneR} matches \textbf{BC} with no distractors ($75.0\%$ vs. $73.3\%$) and outperforms both baselines with three distractors.
    }
    \vspace{-1.0em}
    \label{fig:real-world}
\end{figure*}

\section{Hardware Experiments} \label{sec:experiments}
The controlled simulations of Sections~\ref{sec:sims} and \ref{sec:drone_sim} suggest that InneR outperforms standard behavior cloning and regularization alternatives that penalize the entire spectrum of the inner product matrix.
To evaluate how well InneR performs with physical robots, we present real-world studies across three tasks: ``\textit{closing a coffee maker}'', ``\textit{stacking two cubes}'', and ``\textit{pushing a pot}''.

\begin{figure}
    \centering
    \includegraphics[width=1.0\linewidth]{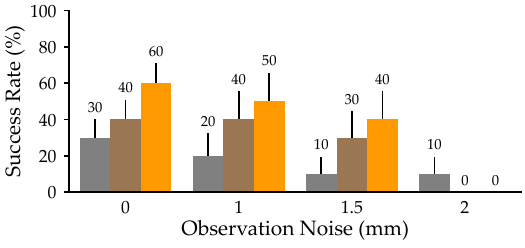}
    \caption{Success rate in \textit{close} with three distractors as Gaussian noise is added to the object positions that the policy observes. The horizontal axis is the standard deviation of the added noise; $0$~mm repeats the three distractor results of \fig{real-world}. \textbf{InneR} has the highest success rate up to $1.5$~mm}
    \label{fig:obs-noise}
    \vspace{-1.0em}
\end{figure}

\begin{figure}
    \centering
    \includegraphics[width=1.0\linewidth]{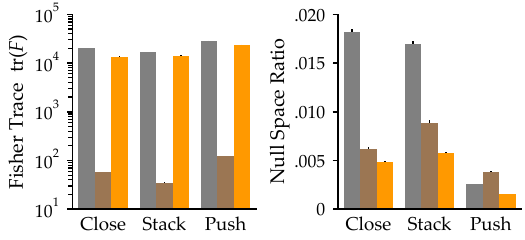}
    \caption{The regularization results from our hardware experiments, measured for the three-distractor diffusion policies at training states. (Left) Trace of the Fisher information $F$ (log scale). (Right) Shared null space ratio of \eq{A5}, with $F$ in place of the inner product, $r=14, k=8$. \textbf{Nuclear} shrinks the Fisher information by more than two orders of magnitude, whereas \textbf{InneR} keeps it close to \textbf{BC}'s level and attains the lowest null space ratio on every task. Error bars show standard error over $8$ batches of $256$ states.}
    \label{fig:hw-null}
    \vspace{-1.0em}
\end{figure}

\p{Experimental Setup}
We use a 7-DoF Franka Emika Panda robot arm with a UMI gripper \cite{chi2024universal} and a stationary Orbbec Femto Mega RGB-D camera.
Objects are segmented with SAM3 \cite{carion2026sam} on the first frame and tracked across frames with EdgeTAM \cite{zhou2025edgetam}.
We project the depth under each mask into the robot's frame of reference to recover each object's pose.
When tracking five objects, this process runs near our control frequency of $10$ Hz on an NVIDIA RTX 3080 GPU.

\p{Training Data}
We collected $25$ teleoperated demonstrations for \textit{close}, $50$ for \textit{stack}, and $25$ for \textit{push}.
To cover more of the workspace, we augmented each set with copies of the demonstrations rotated about the robot's base, so that the arm's motion and every object rotate together. We also mirrored the motions of the robot and the objects about the $y$-axis, for a total of $100$ demonstrations in \textit{push} and $300$ in \textit{stack} and \textit{close}.
The state is the width and pose of the gripper, as well as the pose of each object.
The action is the gripper's target pose and width $0.5~\si{\second}$ ahead.
For each task, the distractor conditions add the position and yaw of an apple, an orange, and a banana to the state. These distractors are irrelevant to the robot's task.

\p{Independent Variables}
We compare three methods from Sections~\ref{sec:sims} and \ref{sec:drone_sim}: \textbf{BC}, \textbf{Nuclear}, and \textbf{InneR}.
Within each task, every method is given the exact same training data, model architecture, and hyperparameters.
The policy is a diffusion model \cite{chi2025diffusion} whose denoiser is an MLP with three hidden layers ($256$ units, ReLU activation), trained for $10^5$ gradient steps using the Adam optimizer ($\alpha=10^{-3}$ with cosine decay). 
We deploy an exponential moving average of the weights (decay $0.999$). Actions are sampled with DDIM \cite{song2020denoising} using $50$ denoising steps.
For \textbf{InneR} we use $M=I$, $k=8$, $\gamma=0.1$, and $r=2m = 14$ ($r=m=7$ with no distractors, as the state has too few dimensions to leave a null space at $r=14$).
We estimate $F$ with $N_a=14$ sampled actions and $N_\varepsilon=4$ draws of $(t, \varepsilon)$ per action.

\p{Dependent Variables}
We measure the success rate over $20$ trials per method, task, and distractor condition.
In \textit{push}, success is when the pot ends in the target region, in \textit{stack}, the black cube rests on top of the tan cube, and in \textit{close}, the coffee maker's lid should be latched closed.
To examine how each regularizer affects the robot's policy, we also report the trace of the Fisher information and the shared null space ratio of \eq{A5}, as in Sections~\ref{sec:sim2} and \ref{sec:drone_sim2}, and each policy's \textit{selectivity}.
We define a policy's \textit{sensitivity} to an object as the average distance that the commanded gripper position moves when the object is displaced by $1$~cm in the table plane (with the diffusion noise held fixed), and its \textit{selectivity} as its mean sensitivity to the task-relevant objects divided by its mean sensitivity to task-irrelevant distractors.

\p{Hypotheses}
We test three hypotheses:
\begin{quote}
    \textbf{H1}: InneR \textit{will outperform behavior cloning without regularization and baselines that regularize the policy globally.}
    \\
    \textbf{H2}: InneR \textit{will be more robust to observation noise than baselines.}
    \\
    \textbf{H3}: InneR \textit{will scale more effectively with the number of distractors than baselines.}
\end{quote}

\p{Results}
Our results are summarized in \fig{real-world}. 
To address \textbf{H1}, we average each method's success rate across the three tasks with zero and three distractors. 
\textbf{InneR} has an average success rate of $65.8\%$, compared to $45.8\%$ for \textbf{BC} and $21.7\%$ for \textbf{Nuclear}.
With no distractors \textbf{InneR} and \textbf{BC} perform similarly ($75.0\%$ and $73.3\%$), so the regularizer does not reduce performance when there is nothing to ignore. 
The gap widens with three distractors: \textbf{InneR} scores $56.7\%$ versus \textbf{BC}'s $18.3\%$.
\textbf{Nuclear} scores worse than \textbf{BC} even without distractors ($25.0\%$) and never succeeds in \textit{stack}.
%

To address \textbf{H2}, we add zero-mean Gaussian noise to every object position that the policy observes in \textit{close} with three distractors, drawn independently at each control step ($10$ trials per method and noise level). 
For reference, the position estimates from our perception pipeline vary by roughly $1$~mm (standard deviation) between frames.
\fig{obs-noise} shows that \textbf{InneR} has the highest success rate with $1$~mm of added noise ($50\%$ vs. $20\%$ for \textbf{BC} and $40\%$ for \textbf{Nuclear}) and with $1.5$~mm ($40\%$ vs. $10\%$ and $30\%$).
At $2$~mm \textbf{Nuclear} and \textbf{InneR} fail. \textbf{InneR} therefore remains the most successful method under moderate observation noise, but its success rate falls about as quickly as the baselines'. 
These results partially support \textbf{H2}.
%

\begin{figure}
    \centering
    \includegraphics[width=1.0\linewidth]{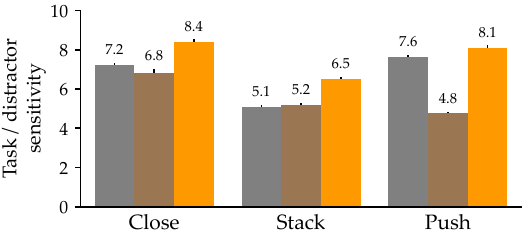}
    \caption{Selectivity of the three-distractor policies on states encountered during the real-world trials (higher is better). \textbf{InneR} is the most selective method on every task. Error bars show standard error over $400$ states.}
    \label{fig:hw-sel}
    \vspace{-1.0em}
\end{figure}

Turning to \textbf{H3}, we consider \textit{close} with zero to three distractors (\fig{real-world}, left).
For each method we fit a logistic regression of trial success on the number of distractors ($80$ trials per method) and apply a Wald test \cite{agresti2002categorical} of the null hypothesis that the regression slope is zero, i.e., that the method's success probability does not change with the number of distractors.
\textbf{BC}'s success rate falls from $75\%$ to $30\%$ (a linear trend of $-14.5$ percentage points per distractor), and the test rejecets the null hypothesis ($z=-2.82$, $p=0.005$).
\textbf{InneR} stays between $60\%$ and $80\%$ ($-5.5$ points per distractor), and its rate of change with respect to the number of distractors is not statistically significant ($z=-1.21$, $p=0.23$).
\textbf{Nuclear}'s rate of change is also not significant ($z=-0.20$, $p=0.84$), but it is the least successful method without distractors.
We next compare the methods when distractors are present by pooling the conditions with one to three distractors ($60$ trials per method).
\textbf{InneR} succeeds in $68\%$ of these trials, whereas \textbf{BC} and \textbf{Nuclear} each succeed in $37\%$.
For each baseline, a two-proportion $z$-test \cite{agresti2002categorical} rejects the null hypothesis that it has the same success probability as \textbf{InneR} ($p < 0.001$ in both cases).
Altogether, these results support \textbf{H3}.

\begin{table}[t]
    \centering
    \caption{Sensitivity of the three-distractor policies}
    \label{tab:hw-sens}
    \begin{tabular}{llccc}
        \hline
        & & \textbf{BC} & \textbf{Nuclear} & \textbf{InneR} \\
        \hline
        \textit{Close} & Task objects & $2.193$ & $0.636$ & $2.072$ \\
                       & Distractors  & $0.305$ & $0.094$ & $0.247$ \\
        \rowcolor{gray!15}
        \textit{Stack} & Task objects & $1.272$ & $0.490$ & $1.378$ \\
        \rowcolor{gray!15}
                       & Distractors  & $0.249$ & $0.094$ & $0.212$ \\
        \textit{Push}  & Task objects & $1.978$ & $2.292$ & $1.712$ \\
                       & Distractors  & $0.259$ & $0.479$ & $0.211$ \\
        \hline
    \end{tabular}
\end{table}

\p{Discussion}
\fig{hw-null} repeats the measurements of Section~\ref{sec:sim2} and \ref{sec:drone_sim2} for the three-distractor policies. 
\textbf{Nuclear} reduces the trace of $F$ by more than two orders of magnitude on every task, whereas \textbf{InneR} retains $66\%$ to $83\%$ of \textbf{BC}'s.
\textbf{InneR} also has the lowest shared null space ratio on all three tasks (e.g., $0.5\%$ vs. $1.8\%$ for \textbf{BC} in \textit{close}).
As in simulation, these results distinguish suppressing the policy's overall sensitivity from regularizing its sensitivity in shared null space directions.

Finally, we investigate how these differences appear in the robot's actions at the states it has encountered during real-world trials (cf. \fig{hw-sel}).
\textbf{InneR} is the most selective method on every task: displacing a task object changes its action $6.5$ to $8.4$ times more than displacing a distractor, compared to $5.1$ to $7.6$ times for \textbf{BC} and $4.8$ to $6.8$ times for \textbf{Nuclear}.
Table~\ref{tab:hw-sens} lists the sensitivities defining each selectivity ratio.
Relative to \textbf{BC}, \textbf{InneR} responds $15\%$ to $19\%$ less to the distractors while its response to the task objects stays within $13\%$ of \textbf{BC}'s.
\textbf{Nuclear}'s effect is task-dependent: in \textit{close} and \textit{stack} it responds $2.6$ to $3.4$ times less than \textbf{BC} to the task objects and the distractors alike, and in \textit{push} it responds $1.8$ times more than \textbf{BC} to the distractors.

\section{Conclusion} \label{sec:conclusion}

We have introduced InneR, a regularization algorithm that makes deterministic or stochastic imitation learning policies robust to local state perturbations.
InneR builds on the insight that nearby states should have consistent first-order responses to the same perturbation.
By clustering states whose inner products are aligned, averaging the state space inner product within each cluster, and then minimizing the smallest eigenvalues of that average, InneR shapes the learned policy to maintain a consistent null space without reducing its expressivity.
Across our experiments, adding InneR to standard behavior cloning made the policy more robust to irrelevant state dimensions without requiring any additional data.
Under distribution shift, InneR roughly halved the final state error in the single-integrator simulations and raised the landing success rate in the drone simulations by up to $17$ percentage points. On hardware, InneR raised the average success rate of an imitation-learning policy with three distractors from $18\%$ to $57\%$, while matching the performance of behavior cloning in tasks with no distractors.

\p{Limitations}
We highlight two related limitations.
First, InneR increases the training time by roughly $5\times$ in our experiments, and this cost grows with state dimension because the decompositions in Algorithm~\ref{alg:1} scale with the size of the inner product matrix.
Second, due to this overhead, we have not explored InneR for policies that input image space observations or high-dimensional feature representations.
Recent research has found efficient ways to estimate other Jacobian norms in high-dimensional spaces \cite{scarvelis2024nuclear}. In future work, we hope to extend these estimates to the eigenvalues of the averaged inner product so that inner product regularization can be used with high-dimensional inputs.

\ifijrr
    \begin{funding}
        This work was supported in part by the National Science Foundation through grants \#2535892, \#2535276, and \#2337884.
    \end{funding}
    \begin{sm}
        Code and videos of our experiments are available on our website: \url{https://collab.me.vt.edu/inner}.
    \end{sm}
    \begin{dci}
        The authors declared no potential conflicts of interest with respect to the research, authorship, and/or publication of this article.
    \end{dci}
\else
    \section*{Funding}

    This work was supported in part by NSF Grant \#2337884.
\fi


\balance
\ifijrr
    \bibliographystyle{SageH}
\else
    \bibliographystyle{IEEEtran}
\fi
\bibliography{Bibtex}

\end{document}